\documentclass[12pt, a4paper]{article}
\usepackage[style=authoryear,backend=bibtex, style=numeric]{biblatex} 
\usepackage{longtable}  
\usepackage{amsmath}
\usepackage{bm}
\usepackage[hidelinks]{hyperref}
\PassOptionsToPackage{hyphens}{url}
\numberwithin{equation}{section} 
\usepackage{subcaption}
\usepackage{graphicx}
\usepackage{float}     
\usepackage{array} 
\usepackage{multirow} 
\usepackage{tabularx} 
\usepackage{caption} 

\begin{document}
\title{DATA SENSOR FUSION IN DIGITAL TWIN TECHNOLOGY FOR ENHANCED CAPABILITIES FOR A HOME ENVIRONMENT}
\author{Benjamin Momoh, Salisu Yahaya}
\date{January, 2025}
\maketitle

\begin{abstract}
  
This paper explores the challenges posed by the coronavirus outbreak and its economic repercussions by investigating the integration of data sensor fusion in digital twin technology to enhance capabilities in a home environment. The study explores the fourth industrial revolution and the role of digital transformation in mitigating disruptions. Central to this investigation is collecting a primary dataset using the Wit Motion sensor, which provides data from accelerometers, gyroscopes, and magnetometers. This sensor was used to collect and analyse data for four distinct activities: walking, working, sitting, and lying, synchronized by timestamp.

By examining the integration of Cyber-physical systems, the Internet of Things (IoT), Artificial Intelligence (AI), and robotics, the project seeks to provide valuable insights into the significance of data fusion in digital twin technologies, particularly within home environments. The study implements and evaluates various sensor fusion methodologies, including feature-level fusion, decision-level fusion, and Kalman filter fusion, to contribute to advancements in AI. Machine learning models such as Support Vector Machines (SVM), Gradient Boosting (GBoost), and Random Forest (RF) are utilized to assess the effectiveness of these fusion techniques. A comprehensive comparison is conducted between individual sensor data from accelerometers, gyroscopes, and magnetic field sensors, and the data obtained through feature-level and decision-level fusion to determine the impact on model performance and to highlight the benefits of sensor fusion in enhancing digital twin capabilities in home environments.

The evaluation of the sensor fusion methodologies used demonstrated an improved performance on the various machine learning models. The results indicate that sensor fusion techniques significantly enhance the accuracy and reliability of the models compared to individual sensor data. Despite the higher accuracy of the magnetometer, its weakness in providing reliable data in certain environments and being prone to external disturbances like metallic and magnetic fields was mitigated through data fusion and improving the overall performance of the models. A close analysis was carried out between the individual sensors' accuracy and the benefits of data fusion using the accuracy score, confusion matrix and their classification outcomes. This approach provided a more complete and nuanced understanding of the activities that, despite an individual sensor performance offering a strong accuracy in ideal conditions, integrating data from multiple sensors ensures greater consistency and reliability in real-world scenarios, making a robust and better system equipped to handle sensor compromise.
\end{abstract}
\section{Introduction}
The Coronavirus outbreak, the first lockdown in the UK on Thursday, March 23, 2020, and the world at large led to an evident and drastic meltdown in economic activities and a setback in productivity. To prevent such effects from repeating themselves shortly, the fourth industrial revolution and digital transformation were the day's focus (\cite{mihai2022}). According to \cite{singh2023}
, the healthcare industry had a noteworthy expansion. This fourth industrial revolution started during the 21st century and has been influenced by recent technological innovations such as Cyber-physical systems, the Internet of Things (IoT), Artificial Intelligence (AI), and robotics, giving rise to the concept of digital twins.  What is a digital twin? The \cite{kerckhove2021} publication briefly defined the digital twin as a "1:1 digital representation of a physical product or process over its whole lifecycle." The digital twin represents the future of technological advancement and sets to persist, transforming how humans operate and perceive their surroundings and transforming imagination into tangible reality.

The practical application of digital twins involves diverse models that continuously enhance data collected from various sensors, ensuring a highly accurate representation of a physical product's lifecycle. The future of digital twins, currently showing an average annual growth rate of about 38\%, is envisioned as a twining process that addresses technological advancements, human history, and biological conditions (\cite{kerckhove2021}). This perspective underscores the transformative potential of digital twins, making it a topic of significant interest and exploration.

Data fusion is employed to enhance the efficacy and effectiveness of digital technology, which can directly impact the level of integration between physical and digital objects. Data fusion uses computer technology to analyse and synthesise information from several sensors obtained according to time series for the required decision-making and estimation task  \cite{he2021data}. Most machine models and AI activities are in cyberspace, and information about building a digital twin, such as the individual's health records, behavioural patterns, and personal preferences, is collected. This information could be exchanged for money.

Ethical considerations in handling and utilising sensitive data from various sources are not just important; they are paramount. Adherence to industry standards and best practices in data fusion and digital twin technology, ensuring data security and privacy measures are robust and compliant with regulations, collaboration and communication challenges among interdisciplinary teams working on the project, addressing potential biases and inaccuracies in data fusion processes and algorithms, and a continuous professional development and skill enhancement to stay up-to-date with evolving technologies and methodologies in the field, are all critical aspects that demand our urgent attention.

The paper is divided into 6 sections; section 2 comprises literature reviews of some well-known data fusion models, fusion classification and architecture. In section 3, we investigated approach and methodology used and the structure of our primary data set. In section 4, Data visualization was carried on our data. Confusion matrix on the dataset with and without fusion was done. In section 5, data analysis evaluation of results was carried out with an explanation of results derived from the SVM, Gradient boost, and the Random Forest used. Discussion on the results was also made. The paper ends up with some concluding remarks, advanced data fusion techniques discussions, and future work plans in section 6.
\section{Literature Review}
This section provides existing literature on the origins of digital twins and their technologies.
Recent years have shown significant developmental progress in digital twins in academia and industry (\cite{wu2020}). The digitization concept has also helped address issues related to improved manufacturing quality, operation objectives, and conditions during the production of machinery energy. The potential of digital twin technology to revolutionize these areas is a cause for optimism. Countries, institutes, and industries worldwide, such as Massachusetts Institute of Technology (MIT) in the United States, Siemens in Germany, and China Nuclear Power Research and Design Institute, have also applied digital twin technology (\cite{mengyan2024digitaltwin}).\\
What is a digital twin? The Cambridge dictionary defines digital as "to record" or "information storage as digits of 1's and 0's, to show the presence of signals", while "Twin," as it is generally known, is one of two things containing or consisting of two matching or corresponding parts linked together. 
Digital Twin started relatively as 3D Computer-Aided Design (CAD) geometry and all design requirements for a product (including notation and parts lists). These Digital Mock-Ups are no longer mere digital copies but now exchange information with their real-life counterparts via a series of attached sensors—making them digital twins. Current experience from domestic and foreign manufacturing industries has made it evident that the product model defined by 3D digitalisation has grown and evolved, and its benefits have been repeatedly verified (\cite{xiong2022}).
\cite{jeong2022} in his paper, defined a digital twin as a replica of "physical objects (e.g., people, objects, spaces, systems, and processes) in the real world into digital objects in the digital world" to address various real-world problems and optimise the natural world through simulation or prediction of situations that can occur in the future. This concept of the digital twin, which was a paradigm shift in technological advancement, was introduced when the National Aeronautics and Space Administration (NASA) decided to create the physical twin of a space aircraft within the Apollo Program to reproduce its behaviour in space (\cite{wang2020digital}). The work of Michael Grieves with John Vickers of NASA presented the concept of the product lifecycle, which is from the physical product, a virtual representation of that product, and the bi-directional data connection that feeds data from the physical to the virtual and vice versa (\cite{jones2020}), (\cite{macias2024}). The digital twin concept refers to a digital twin instead of a physical one, composed of a physical part, a digital part, and interconnectivity for data transfers. 
The digital twin requires high integration between the digital and physical parts through data transfer. The level of integration consists of a digital model representing a physical entity, a digital shadow representing a uni-directional inflow of information (Physical-to-Virtual), and a digital twin representing a bi-directional inflow of information (Virtual to physical). This integration is possible by using sensors installed in the physical object parts to reflect the digital objects. Similarly, the digital object can change the physical state through these sensors and actuators. According to \cite{kritzinger2018}, the level of integration of a digital twin with its physical counterpart has all to do with the level of data integration, which he also identified in three levels: the digital model, digital shadow, and the digital twin. 
\subsection{Digital Twin Technologies}
For a digital twin to exist, data must flow in and out between the digital and physical objects in real-time. Sensors and actuators make this flow possible, and various technologies collect and store data in real time. \cite{attaran2023}, gave four different types of technologies. These are The Internet of Things (IoT), Artificial Intelligence (AI), Extended Reality (XR), and the Cloud. A particular technology depends on the digital twin use case, including industrial public and personal areas. These technologies can used for insightful information in visualisation and operation technology, analysis technology, multidimensional modeling and simulation technology, connection technology, data and security technology, and synchronisation technology. 

\subsection{Data Fusion}
Data fusion, also known as information fusion, is a process that combines data or information from various sources to enhanced decision-making. In the context of data fusion architectures, the terms' information,' 'data,' and 'knowledge' describe the hierarchical order levels of a data fusion process. Whether referred to as information fusion or data fusion, the primary focus is on the fundamental questions related to fusion. These questions include the fusion's objectives, the pieces of information or data to be fused, their characteristics and level of uncertainty, the available fusion methods, and the associated difficulties and challenges.  
In Digital Twins, data fusion and information are used interchangeably. Data Fusion, in this context, is a Crucial process. It combines data from multiple sensors to accurately and reliably represent the physical and digital systems. This enhanced representation is pivotal in improving decision-making, underscoring the significance of data fusion in the digital twin domain and its potential impact on your work.  
Data fusion is a multidisciplinary research area that draws ideas from various fields. It is defined as the study of efficient methods for transforming information from different sources and points in time into a representation that effectively supports humans or automated quality. With data fusion in the digital twin, we can produce a refined digital representation of our physical object with characteristics that will enhance decision-making and control-related activities such as environmental mapping, object recognition, forecasting, and prevention. Data fusion is familiar and has been used since the 1960s, notably by the US Department of Defense and the Joint Directors of Laboratories (JDL). Data fusion has applications in diverse fields, such as robotics, defence, and healthcare. However, the application of data fusion in digital twins is relatively new and holds great potential to enhance the accuracy of a digital twin.  
Direct fusion involves combining sensor data from heterogeneous or homogeneous sources. In contrast, Indirect fusion combines the outputs of heterogeneous information deduced from sensor data. Some researchers often interchange 'data fusion' with 'information fusion.' 
Direct data fusion, categorised as low-level fusion, involves combining sensor data from heterogeneous or homogeneous sources. Indirect data fusion, on the other hand, is classified as high-level fusion since it is performed after some analysis. There is also mixed data fusion, low-level and high-level fusion. Understanding these different levels of data fusion can make you feel more informed and knowledgeable about the topic.
\begin{table}[htbp]
\caption{Types of data fusion and their level of fusion.\label{tab1}}
\begin{tabularx}{\textwidth}{|l|l|X|l|}
\hline
\textbf{S/N} & \textbf{DATA FUSION TYPES} & \textbf{STAGES OF FUSION} & \textbf{Level of Fusion} \\
\hline
1 & Direct Fusion & Fusion of data from sensors before analysis & Low-level fusion \\
2 & Indirect Fusion & Fusion after analysis & High-level fusion \\
\hline
\multirow{2}{*}{3} & \multirow{2}{*}{Complex Fusion} & \raggedright Fusion of multiple sensor data and advanced processing & \multirow{2}{*}{High-level fusion} \\
& & \raggedright (e.g., Kalman Filter, Bayesian networks) & \\
\hline
\end{tabularx}
\end{table}
\subsection{Methods And Techniques of Data Fusion}
Data fusion's main objective is to increase the reliability of the decisions made using collected data from sensors. Researchers generally categorise data fusion methods into four main methods. 	  
Probabilistic method, Statistical method, Knowledge base theory method and  Reasoning method. The probabilistic Data fusion method, which follows Bayes' Rule, is the heart of most data fusion methods. The Probabilistic Data Fusion method has practical applications in establishing a joint probability distribution P(y,z) between the relationship y and z for discrete and continuous variables. This practical relevance makes it a crucial method for understanding data fusion. 
Statistical methods include the cross-covariance intersection. Knowledge base theory, a widely popular method for handling uncertainty in data fusion, includes intelligence aggregation methods, such as Fuzzy logic. Evidence reasoning, often called the Dempster-Shafter evidence theory, is used in areas related to automated reasoning applications and recursive operators.  
The models we shall analyse in this research will be based on some essential characteristics of the data and the type of architecture. 

\subsection{Relation Between Input Data Sources}
Data generated from different sources could have relationships with each other about a target. Durrant-Whyte proposes that these relationships be defined as complementary, redundant, or cooperative.\\ 
The information provided by the input sources represents different parts of the scene and could thus be used to obtain generally accepted information. Such data types are complementary in nature. For instance, when two separate sensors provide information on a particular target.\\
Data that can be overlapped or superimposed on each other are redundant. Fusing redundant data can improve target confidence. Redundant data are mainly observed when two or more sources provide individual information on a particular target.\\
Cooperative Data combines provided information with new data, a key concept in data fusion. This process typically results in more complex information than the original data, such as \textbf{multi-modal (audio and video) data fusion}.\\
The type of Data input/output is a classification system that breaks data fusion into five categories based on their nature.\\
Data in - Data out (DAI-DAO):
This fusion process inputs data to make them more accurate. It is the most elementary data fusion method in classification. This data fusion process inputs and outputs raw data, but the results are typically more reliable or accurate.\\

Data in - Feature out (DAI-FEO):
At this level, the data fusion process employs raw data from the source inputs to extract features or characteristics that describe an entity in the environment.\\
Feature in - Feature out (FEI-FEO):
    In FEI-FEO, the inputs and outputs of the data fusion process are features. This data fusion process addresses feature improvement and is regarded as information fusion.\\
Feature in - Decision out (FEI-DEO):Most algorithms fall into this category because of their feature purpose classification. FEI-FEO obtains a set of features as input and provides a set of decisions as output.
Decision in - Decision Out (DEI-DEO): This is the highest level in this classification. DEI-FEO fusion transforms some decisions at the low level into global decisions for decision-making. It fuses input decisions to obtain better or new choices. 

\subsection{JDL Data Fusion Classification}
\begin{figure}[htpb]  
   \centering
    \includegraphics[width=\textwidth]{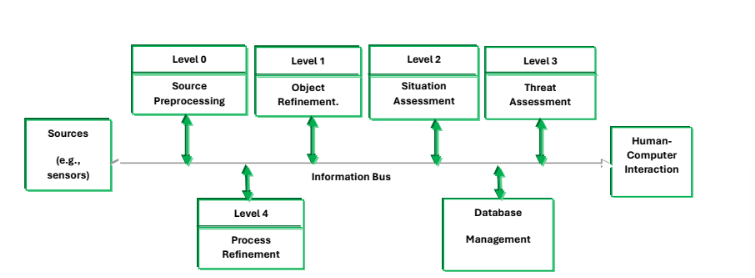}  
    \caption{JDL Architecture}
    \label{fig:datafusionclassification}
\end{figure}
This classification is the most famous conceptual model in the data fusion community. It was initially proposed by JDL and the American Department of Defense (DoD) for use in the military. These organisations classified the data fusion process into five processing levels.\\ 
Level 0- (Sources Preprocessing) provides the input data (lowest level). Different sources, such as sensors, a priori information (references or geographic data), databases, and human inputs, can be employed. The primary aim of level 0 is data transformation and transfer to the proper level for further processing.\\ 
Level 1-(Object Refinement) employs the processed data from the previous level. The main aim is to identify entities and relations. Standard procedures at this level include spatiotemporal alignment, association, correlation, clustering or grouping techniques, state estimation, the removal of false positives, identity fusion, and combining features extracted from images. The output results of this stage are object discrimination (classification and identification) and object tracking (state of the object and orientation). This stage transforms the input information into consistent data structures.\\
Level 2 focuses on a higher level of inference than Level 1. The relationship information gained from the previous level broadens the scope of the investigation into the entire environment of the entity. Situation assessment aims to identify the likely situations given the observed events and obtained data. It establishes relationships between the objects. Relations (i.e., proximity and communication) are valued to determine the significance of the entities or objects in a specific environment. The aim of this level includes performing high level inferences and identifying significant activities and events (patterns in general). The output is a set of high-level inferences.     
Level 3 evaluates the impact and threats of the detected activities in Level 2 to obtain a proper perspective. The current situation is assessed by predicting a logical outcome's risks, vulnerabilities, and probabilities of operation. 
Level 4 -(Process Refinement), improves the process from level 0 to level 3, a managed part of the entire process. The aim is to achieve efficient resource management by monitoring other levels in real-time while accounting for task priorities, scheduling, and controlling available resources. The supporting components of the JDL architecture are;
Human-Computer Interaction (HCI), 
Database Management System, 
 and Sources, which is the base of the whole system.\\ 
 
One of the limitations of the JDL method is how the uncertainty about previous or subsequent results could be employed to enhance the fusion process (feedback loop). \cite{liggins2009handbook}, propose several refinements and extensions to the JDL model. \cite{blasch2010high} proposed to add a new level (user refinement) to support a human user in the data fusion loop. The JDL model represents the first effort to provide a detailed model and a common terminology for the data fusion domain. However, because their roots originate in the military domain, the employed terms are oriented to the risks commonly occurring in these scenarios.  
The Dasarathy model differs from the JDL model in terms of the terminology and approach employed. The former is oriented toward the differences between the input and output results, independent of the employed fusion method. The Dasarathy model provides a method for understanding the relations between the fusion tasks and employed data. In contrast, the JDL model presents an appropriate fusion perspective for designing data fusion systems. 
\subsection{Classification Based On The Type Of Architecture}
\begin{figure}[h!]
    \centering
    \includegraphics[width=\linewidth]{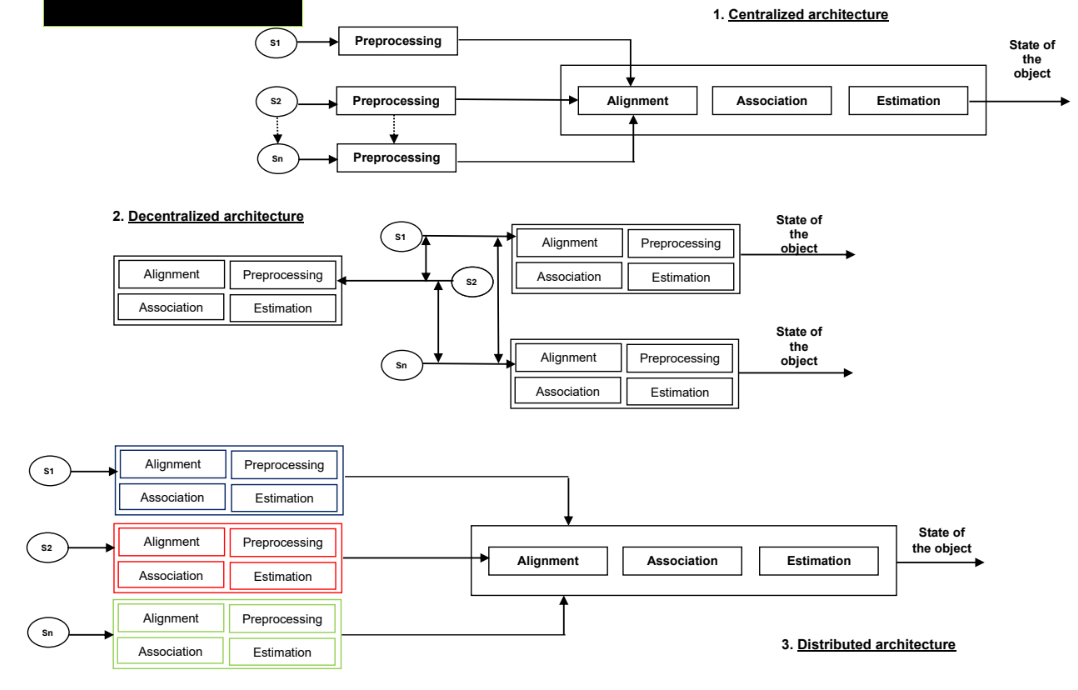}  
    \caption{Types of Data Fusion Architectures}
    \label{fig:landscapeimage}
\end{figure}
One of the main questions when designing a data fusion system is where the data fusion process will be performed. Based on this criterion, the following types of architectures could be identified by research
\begin{enumerate}
\item\textbf{Centralised Architecture} \\In a centralised architecture, the fusion node resides in the central processor that receives the information from all the input sources, which implies that all the fusion processes will be executed in a central processor. In this schema, the sources obtain only the observational measurements and transmit them to a central processor, where the data fusion process is performed. If data alignment and association are performed correctly and the required data transfer time is insignificant, then the centralised scheme can provide significant benefits with its theoretical optimality.

The disadvantage of this kind of architecture is that time synchronisation with the various sensors is a significant challenge in real time. Additionally, the bandwidth cost of transferring data to a central processor is expensive and can lead to information loss, highlighting the need for further research and development in these areas.
\item\textbf{Decentralised Architecture} \\A decentralised architecture comprises a network of nodes. Each node has its processing capabilities, implying no single data fusion point. Therefore, for the fusion process, each node uses its local information in conjunction with the information received from its peers. Data fusion is performed autonomously, with each node accounting for its local information and the information received from its peers. Thus, this type of architecture could suffer scalability problems when the number of nodes increases. 
\item\textbf{Distributed architecture} \\In a distributed architecture, each generated data is analysed independently from the local node before the information is sent to the fusion node. Machine learning model methods like the K-NN, Multiple Hypothesis Testing (MHT), and Probabilistic Data Association (PDA) are methods used to associate and estimate at the source nodes. In other words, data association and state estimations are done only with their local analysis, and the analysed information becomes input for the fusion process, providing a fused global analysis.
\item\textbf{Hierarchical architecture}
\\This architecture combines decentralised and distributed nodes, creating hierarchical schemes where the data fusion process is executed at different levels in the hierarchy. Implementing a decentralised data fusion system can be challenging due to the significant computation and communication requirements. However, it is essential to note that there is no single best architecture existence yet known.  
According to research and studies, selecting the most appropriate architecture should be based on the requirements, existing networks, if there are any, data availability, node processing capabilities or hardware, and the organisation of the data fusion system.  
According to \cite{castanedo2013}, the methods for data fusion are classified into three basic categories which are:
\begin{itemize}  
    \item\textbf{Data Association} \\Data association is a complex task that aims to establish the set of observations/measurements generated by the same target over time. It involves intricate methods such as the K-NN, Probabilistic Data Association (PDA), Joint Probabilistic Data Association (JPDA), and Multiple Hypothesis Test (MHT).    
    \item\textbf{State Estimation} \\State estimation is a precise process that considers the state of a target under movement (i.e. position), given the observation or measurement. It relies on accurate methods such as Maximum Likelihood Estimation (MLE), Kalman Filter, Distributed Filter, and covariance consistency methods. 
    \item\textbf{Decision Fusion} \\Decision fusion is crucial in making high-level inferences about an event and its activities. It does so by analyzing the detected targets provided by many sources, highlighting their importance.
    \end{itemize}
\end{enumerate}

\subsection{Data Fusion Challenges}
We must implicitly examine the methods outlined above, which try to address challenges in data fusion. 
Understanding data imperfection is crucial as it forms the basis of all data fusion methods. Sensor data, often imprecise, uncertain, ambiguous, vague, and incomplete, presents a significant challenge. Although we can improve data quality by modeling its imperfection and using other available information and powerful mathematical tools, the severity of data imperfection can significantly affect fusion quality if data fusion fails to extract precise and valuable data (\cite{khaleghi2013multisensor}).\\

Some data uncertainties are caused by inherent noise in measurement, sensors, and environments. These noises lead to data outliers or disorder, collectively called data inconsistency.  Data inconsistency introduces a terrible effect on data fusion if the fusion model cannot distinguish the techniques necessary to overcome this problem by eliminating the influence it creates. There are some other problems caused by lasting or dynamic failures, which are challenging to model and predict in the usual way (\cite{bakr2017distributed}), (\cite{khaleghi2013multisensor}).\\

Data-related challenges often appear in a system that applies belief functions or Dempster-Shafer theory \cite{meng2020survey} when some problems that should be treated independently are erroneously integrated.\\

 Data captured from different sensors with different frames must undergo a crucial step before fusion alignment into a standard frame. This process, known as data alignment, is of utmost importance in data fusion. If the data fusion algorithm fails to address this, the lack of alignment or correlation can lead to over/under confidence or biased estimation, underscoring the necessity of this step (\cite{meng2020survey}).\\
 
Data captured by different sensors could have different structures. The data fusion method should be able to integrate the different types to describe the environment better.\\

Another challenge is that of Fusion Location. To determine the fusion location, we must consider a trade-off between fusion cost and quality. This is a problem with respect to wireless sensors. When data are generated from a central node, the expense is bandwidth and time, but if data are from a local node, the cost is data accuracy.\\

In addition, the complexity of data fusion is caused by timeliness, where the significance of a data point might be limited to a limited period, especially for a time-varying system. The fusion node must distinguish the correct order of the data and its validation.

\begin{table}[H]
\centering
\caption{General Types Of Sensors For Data Acquisition Within A Home Environment.}
\begin{tabular}{|p{3.5cm}|p{5cm}|p{5cm}|} 
\hline
\textbf{CATEGORIES} & \textbf{EXTERNAL SENSORS} & \textbf{MOBILE SENSORS} \\
\hline
\parbox[t]{4.5cm}{ENVIRONMENTAL \\ SENSORS} & Barometer, Humidity, Light sensor, Thermal sensors. & Ambient air temperature and pressure, Barometer, Photometer, Thermal sensor. \\
\hline
\parbox[t]{4.5cm}{LOCATION \\ SENSORS} & GPS receiver, Wi-Fi. & GPS receiver, Wi-Fi location. \\
\hline
\parbox[t]{4.5cm}{MOTION \\ SENSORS} & Accelerometer, Pressure sensor, Gravity sensor, Rotational sensor. & Accelerometer, Orientation sensor, Gravity sensor. \\
\hline
\parbox[t]{4.5cm}{IMAGE AND \\ VIDEO \\ SENSORS} & Digital camera, 3D camera, Optical sensor, Infrared sensor & Digital camera, Infrared sensor. \\
\hline
\parbox[t]{4.5cm}{PROXIMITY \\ SENSORS} & Touch sensor, Proximity sensor, RFID, Tactile sensor & RFID, Touch sensor, Proximity sensor. \\
\hline
\parbox[t]{4.5cm}{ACOUSTIC \\ SENSORS} & Microphone, Silicon wave device, Silicon microphone. & Microphone. \\
\hline
\parbox[t]{4.5cm}{MEDICAL \\ SENSORS} & Blood pressure, Dosage control, Stress sensor, Heart rate sensor, Electrodermal activity sensor. & - \\
\hline
\parbox[t]{4.5cm}{CHEMICAL \\ SENSORS} & Oxygen saturation, Electrochemical gas sensor, Aroma sensor. & - \\
\hline
\parbox[t]{4.5cm}{OPTICAL \\ SENSORS} & Fibre optic sensors, Infrared sensor, Radio frequency sensor. & Infrared sensors, Radio frequency sensors. \\
\hline
\parbox[t]{4.5cm}{FORCE \\ SENSORS} & Force-sensitive resistor, Mass sensor, Fingerprint sensor. & Fingerprint sensor. \\
\hline
\end{tabular}
\end{table}
\subsection{Fusion Quality Assessment}
 Okolie and Smit, (2022), proposed three approaches that are commonly used for Digital fusion estimation Models.
\begin{enumerate}
    \item{A qualitative approach which involves an inspection and comparison of results from the physical entity with the digital twin.}
    \item{The quantitative approach, a practical and widely used method, involves using statistical metrics to measure the relative and absolute quality of the fused Digital Elevation Model (DEM).}
    \item{The performance-based approach offers a versatile set of quality assessment criteria, including the Mean Absolute Error (MAE), Coefficient of Determination \( R^{2} \) and Model Loss, allowing for a comprehensive evaluation of a fused digital twin model.}
\end{enumerate}
\section{Approach And Methodology}
This section outlines the methodology for evaluating the impact of data fusion techniques on the performance of a home environmental activity recognition model. Our objective is to demonstrate how combining sensor data from different sources can enhance the capability of machine learning models in classifying activities. The Wit Motion sensor, a crucial tool that provides data from the accelerometer, gyroscope, and magnetometer, will collect and analyze data for four distinct activities: walking, working, sitting, and lying, with similarities only on timestamp, for synchronization.
\subsection{Data Collection}
The Wit Motion sensor will be used to gather data on three different types of sensors:
\begin{itemize}
    \item{Accelerometer} Measures acceleration along the X, Y, and Z axes.
    \item{Gyroscope} Measures angular velocity along the X, Y, and Z axes.
    \item{Magnetometer} Measures the magnetic field strength along the X, Y, and Z axes.
\end{itemize}
Each data point will include a timestamp for time-based analysis and synchronization across different sensors. The data will be collected for four activities: Walking, Working, Sitting, and Lying.
\subsection{Data Structure} The collected dataset will be organized into a pandas DataFrame with the following structure.
\begin{table}[H]
\centering
\caption{Sensors Used And Types Of Data Collected.}
\begin{tabular}{|l|l|}
\hline
\textbf{Columns Names (features)} & \textbf{Types} \\
\hline
Timestamp & Time of the data collection \\
\hline
Acceleration X (g) & Acceleration in the X direction \\
\hline
Acceleration Y (g) & Acceleration in the Y direction \\
\hline
Acceleration Z (g) & Acceleration in the Z direction \\
\hline
Angular velocity X (°/s) & Angular velocity in the X direction \\
\hline
Angular velocity Y (°/s) & Angular velocity in the Y direction \\
\hline
Angular velocity Z (°/s) & Angular velocity in the Z direction \\
\hline
Magnetic field X ($\bm{B_x}$) & Magnetic field strength in the X direction \\
\hline
Magnetic field Y ($\bm{B_y}$) & Magnetic field strength in the Y direction \\
\hline
Magnetic field Z ($\bm{B_z}$) & Magnetic field strength in the Z direction \\
\hline
Target variable indicating the activity class &  walking, working, sitting, and lying \\
\hline
\end{tabular}
\end{table}
The proposed methodology includes several stages, each evaluating different data fusion techniques and their impact on classification performance.
\subsection{Individual Sensor Models}
\begin{itemize}
    \item{Data Preparation:} Each sample in our dataset consists of 20 features obtained from the accelerometer, gyroscope and magnetometer sensors. A total of 3,239 samples were used for this study.
Split the data into training and testing sets (80\% training, 20\% testing) for each sensor type.
    \item{Model Training:} Train separate machine-learning models for each sensor using the following algorithms:
    \begin{enumerate}
        \item{Random Forest (RF):}An ensemble learning method based on multiple decision trees. Each Random Forest (RF) is an ensemble learning method that combines multiple decision trees to improve the accuracy of our activity recognition classification. For a RF, each decision tree in the forest is trained on a random subset of the data (in this case, 80\% of the total data set). 
        
        If \( \mathbf{X} \) represents our input feature vectors (accelerometer, gyroscope, and magnetometer), the prediction variable \( \hat{y}_n \) from the \( n \)-th tree is given by
\begin{equation}
\hat{y}_n = T_n (\mathbf{X})
\label{eq:1.0}
\end{equation}

where \( T_n \) represents the decision function of the tree, and \( n \) is the number of trees. A majority vote of all trees obtains the final decision. That is,
\begin{equation}
\hat{y} = \text{majority\_vote}\left( \{ T_n (\mathbf{X}) \}_{n=1}^{100} \right)
\label{eq:1.1}
\end{equation}

The random forest method reduces overfitting by averaging multiple trees built from different parts of our dataset.

        \item{Support Vector Machine (SVM):} ): A supervised learning model used for classification designed to find the hyperplane that best separates our classes in the feature space, maximizing the margin between them.
 
For our set of labelled training data \( \{ ( \mathbf{X}_i, y_i ) \}_{i=1}^{N} \), where \( y_i \in \{-1, 1\} \), and regarding our dataset for the activity recognition and classification problem with four different classes, each classifier (one for each of the four classes, say class \( k \)) is trained to distinguish between the class \( k \) and all other classes. It implies that for this work:

\[
y_i = \begin{cases}
1 & \text{if the class belongs to class } k, \\
-1 & \text{if the class belongs to any of the remaining three classes.}
\end{cases}
\]

In this scenario, our four (4) SVM training models would be:

\begin{itemize}
    \item Model 1: Distinguishes between "Walking" (\(y_i = 1\)) and not "Walking" (\(y_i = -1\)).
    \item Model 2: Distinguishes between "Working" (\(y_i = 1\)) and not "Working" (\(y_i = -1\)).
    \item Model 3: Distinguishes between "Sitting" (\(y_i = 1\)) and not "Sitting" (\(y_i = -1\)).
    \item Model 4: Distinguishes between "Lying" (\(y_i = 1\)) and not "Lying" (\(y_i = -1\)).
\end{itemize}

The prediction of SVM is based on the class with the highest confidence score or the classifier with the highest output value. This approach is popularly known as the One-vs-Rest (OvR) approach because each SVM will be a binary classifier. However, these models distinguishing themselves from the rest of the model classes work together to perform multi-class classification.

For each class \( k \), the decision function is expressed as:

\begin{equation}
f_k (\mathbf{X}) = \mathbf{w}_k \cdot \mathbf{X} + b_k,
\label{eq:1.2}
\end{equation}

where:
\begin{itemize}
    \item \( \mathbf{w}_k \) is the weight vector for class \( k \),
    \item \( b_k \) is the bias term for class \( k \).
\end{itemize}

The final predicted class for an input \( \mathbf{X} \) is given by:

\begin{equation}
\hat{y} = \arg \max_{k} \left( f_k (\mathbf{X}) \right).
\label{eq:1.3}
\end{equation}

        \item{Gradient Boosting (GB):} A boosting technique builds models sequentially to correct errors of previous models for each class of our dataset.

Let \( \mathbf{X} \) be our input feature vector from acceleration \( \left( A_x, A_y, A_z \right) \), angular velocity \( \left( \omega_x, \omega_y, \omega_z \right) \), and magnetic field \( \left( M_x, M_y, M_z \right) \), obtained from the accelerometer, gyroscope, and magnetometer sensors, respectively. Therefore,
\begin{equation}
\mathbf{X} = \left[ A_x, A_y, A_z, \omega_x, \omega_y, \omega_z, M_x, M_y, M_z \right].
\label{eq:1.4}
\end{equation}

With the actual class label \( y \), where \( y \in \{ 0, 1, 2, 3 \} \), representing walking, working, sitting, and lying, respectively, the Gradient Boost model predicts a score \( F_k (\mathbf{X}) \) for each class \( k \in \{ 0, 1, 2, 3 \} \).

For each class \( k \), the probability of that activity is given by:
\begin{equation}
\ P_k (\mathbf{X}) = \frac{\exp(F_k (\mathbf{X}))}{\sum_{j=0}^{3} \exp(F_j (\mathbf{X}))} = 1.
\label{eq:1.5}
\end{equation}
For the Gradient Boost model, the cross-entropy function across the four classes is minimized as shown below:
\begin{equation}
L = - \sum_{i=1}^{N} \sum_{k=0}^{3} 1 \{ y_i = k \} \log P_k (\mathbf{X}_i),
\label{eq:1.6}
\end{equation}
where:
\begin{itemize}
    \item \( N \) is the total number of training samples.
    \item \( 1 \{ y_i = k \} \) is an indicator function that is 1 if the actual class label for the \( i \)-th sample is \( k \), and 0 otherwise.
    \item \( P_k (\mathbf{X}) \) is the predicted probability for class \( k \) for the \( i \)-th sample.
\end{itemize}

Boosting at stage \( m+1 \) is done after stage \( m \) by updating the score for each class \( k \):
\begin{equation}
F_{k, m+1} (\mathbf{X}) = F_{k, m} (\mathbf{X}) + \eta \cdot h_{k, m} (\mathbf{X}),
\label{eq:1.7}
\end{equation}
where \( h_{k, m} (\mathbf{X}) \) is a new model trained to predict the residual for class \( k \) at stage \( m \).

This residual for each class \( k \) and model stage \( m \) represents the difference between the true class label and the predicted probability. Mathematically, it is expressed as:
\begin{equation}
r_{ikm} = 1 \{ y_i = k \} - P_{k, m} (\mathbf{X}_i).
\label{eq:1.8}
\end{equation}
Here, \( \eta \) is the learning rate that controls how much each model influences the overall prediction.

Hence, the final predicted activity class \( \hat{y} \) for an input \( \mathbf{X} \), derived from the sensor data, is determined by:
\begin{equation}
\hat{y} = \arg \max_k \left( F_k (\mathbf{X}) \right).
\label{eq:1.9}
\end{equation}
    \end{enumerate}

\item{Model Evaluation:} Evaluate each model's performance using accuracy scores and confusion matrices. Their Root Mean squared Error was also used for evaluation.
Model evaluation is done on the accuracy without data fusion on individual sensors based on the classification of activity recognition, and also evaluation on data fusion (feature-level and decision-level fusion).
\subsection{Feature-Level Fusion} 
\begin{enumerate}
    \item{Data Preparation:} Feature-level fusion involves the combination or fusion of the data acquired from the sensors before any decision is taken. We combined features from accelerometer, gyroscope, and magnetometer into a single feature set such that:
    Combined\_dataset = acceleration + angular-velocity + magnetic-field.
    This combination is made possible by timestamp synchronization. Each sensor's data was collected simultaneously, enabling the dataset to be combined on the same label classes for activity recognition.
    \item{Model Training:} The training was conducted using the training dataset with Random Forest, SVM, and G-Boost models on individual sensor data and the combined feature set (feature fusion and decision-level fusion).
    \item{Model Evaluation:}A crucial step in the machine learning pipeline is assessing the model's performance using accuracy scores and confusion matrices. The accuracy score provides a ratio of correctly predicted instances to the total cases. With the confusion matrix, we can have a breakdown of the models’ performance indicating the number of true positives, true negatives, false positives and false negatives classes. Also, The Root Mean Square Error (RMSE) is a metric tool for evaluating the error between the predicted class probability and the actual class labels. Unlike the accuracy score and the confusion matrix that provides a qualitative measure of our model performance, RMSE provides a quantitative measure of prediction error, with a lower RMSE indicating that model predictions are closer to the actual values.

\end{enumerate}
\subsection{Decision-Level Fusion}
\begin{enumerate}
    \item{Model Training:}Train individual models for each sensor as described above.
    \item{Prediction Aggregation:}
    Use majority voting to combine predictions from individual models.
    \item{Majority Voting:}
    Each model votes for a class, and the class with the most votes is chosen as the final prediction.
    This was also carried out with our choices of models.
    \item{Model Evaluation:}
    Evaluate the decision fusion approach using accuracy scores and confusion matrices.
\end{enumerate}
\subsection{Kalmer Filter Equations}The general Kalman filter consists of prediction and update steps. Let's define the variables and equations step by step.

\subsubsection*{Prediction Step}

\textbf{Predicted State Estimate:}

The state vector at time \( k \), given the previous state estimate at time \( k-1 \), is predicted as:

\begin{equation}
\hat{x}_{k|k-1} = F \hat{x}_{k-1|k-1} + B u_k
\label{eq:1.10}
\end{equation}

Where:
\begin{itemize}
    \item \( F \) is the state transition matrix, which in this case is the identity matrix \( I \) (i.e., no change in state unless updated by measurements).
    \item \( \hat{x}_{k-1|k-1} \) is the previous state estimate.
    \item \( u_k \) is the control input, assumed to be zero in this case (no external influence).
    \item \( B \) is the control matrix, which is also zero.
\end{itemize}

Since \( F = I \) and \( u_k = 0 \), the prediction simplifies to:

\begin{equation}
\hat{x}_{k|k-1} = \hat{x}_{k-1|k-1}
\label{eq:1.11}
\end{equation}

\textbf{Predicted Covariance:}

The predicted covariance matrix at time \( k \) is given by:

\begin{equation}
P_{k|k-1} = F P_{k-1|k-1} F^T + Q
\label{eq:1.12}
\end{equation}

Where:
\begin{itemize}
    \item \( P_{k-1|k-1} \) is the previous covariance estimate.
    \item \( Q \) is the process noise covariance matrix (set to \( 0.1I \) in this case).
\end{itemize}

Since \( F = I \), the covariance prediction simplifies to:
\begin{equation}
P_{k|k-1} = P_{k-1|k-1} + Q
\label{eq:1.13}
\end{equation}

\subsection*{Update Step}

\textbf{Kalman Gain:}

The Kalman gain \( K_k \) is computed as:

\begin{equation}
K_k = P_{k|k-1} H^T \left( H P_{k|k-1} H^T + R \right)^{-1}
\label{eq:1.14}
\end{equation}

Where:
\begin{itemize}
    \item \( P_{k|k-1} \) is the predicted covariance matrix.
    \item \( H \) is the measurement matrix (mapping the state to the measurement space).
    \item \( R \) is the measurement noise covariance matrix (set to \( 0.5I \)).
\end{itemize}

\textbf{State Update:}

The updated state estimate is given by:

\begin{equation}
\hat{x}_{k|k} = \hat{x}_{k|k-1} + K_k \left( z_k - H \hat{x}_{k|k-1} \right)
\label{eq:1.15}
\end{equation}

Where:
\begin{itemize}
    \item \( z_k \) is the actual measurement at time \( k \).
    \item \( \hat{x}_{k|k-1} \) is the predicted state.
    \item \( H \hat{x}_{k|k-1} \) is the predicted measurement.
    \item \( z_k - H \hat{x}_{k|k-1} \) is the measurement residual (or innovation).
\end{itemize}

\textbf{Covariance Update:}

The updated covariance matrix is computed as:

\begin{equation}
P_{k|k} = \left( I - K_k H \right) P_{k|k-1}
\label{eq:1.16}
\end{equation}

Where:
\begin{itemize}
    \item \( I \) is the identity matrix.
    \item \( K_k H \) adjusts the covariance based on the Kalman gain and measurement matrix.
\end{itemize}

For our methodology, the state vector,\( \hat{x}_{k-1|k-1} \), includes the filtered \( X \), \( Y \), and \( Z \) values for the Kalman filter.These values are the core components of the system's state, which we aim to estimate and update using the Kalman filter process.

The state vector is represented as:

\begin{equation}
\hat{x}_{k-1|k-1} = 
\begin{bmatrix}
\hat{x}_{k-1|k-1}^X \\
\hat{x}_{k-1|k-1}^Y \\
\hat{x}_{k-1|k-1}^Z
\end{bmatrix}
\label{eq:1.17}
\end{equation}

Where:
\begin{itemize}
    \item \( \hat{x}_{k-1|k-1}^X \), \( \hat{x}_{k-1|k-1}^Y \), and \( \hat{x}_{k-1|k-1}^Z \) represent the filtered values of the \( X \), \( Y \), and \( Z \) components of the state vector, respectively, at time \( k-1 \).
\end{itemize}

The observations used in the update step come from the combined data from the following sensors:
\begin{itemize}
    \item \textbf{Accelerometer}: Provides measurements of acceleration along the \( X \), \( Y \), and \( Z \) axes.
    \item \textbf{Gyroscope}: Provides angular velocity measurements along the \( X \), \( Y \), and \( Z \) axes.
    \item \textbf{Magnetometer}: Provides magnetic field measurements along the \( X \), \( Y \), and \( Z \) axes.
\end{itemize}

These measurements are combined into a vector \( z_k \) at each time step \( k \), which is used in the update step of the Kalman filter to refine the state estimate. The measurement vector \( z_k \) is represented as:

\begin{equation}
z_k = 
\begin{bmatrix}
z_k^X \\
z_k^Y \\
z_k^Z
\end{bmatrix}
\label{eq:1.18}
\end{equation}

Where:
\begin{itemize}
    \item \( z_k^X \), \( z_k^Y \), and \( z_k^Z \) are the measurements at time \( k \) from the accelerometer, gyroscope, and magnetometer along the respective axes.
\end{itemize}

Thus, the state vector \( \hat{x}_{k-1|k-1} \) is updated based on these combined sensor measurements, improving the estimate of the system's state over time.
\end{itemize}
We applied the Kalman Filter to fuse measurements from the accelerometer, gyroscope, and magnetometer across all data points to produce the “Kalman Filtered \(X\),” “Kalman Filtered \(Y\),” and “Kalman Filtered \(Z\).” These values would be stored in a DataFrame and used for estimation.

\section{Implementaion}All code and implementation were done using Python due to its extensive libraries and tools for data analysis and machine learning. Python's simplicity and readability also make it a popular choice for data science projects.
\subsection{Data Visualization}
\begin{figure}[h!]
    \centering
    \includegraphics[width=\linewidth]{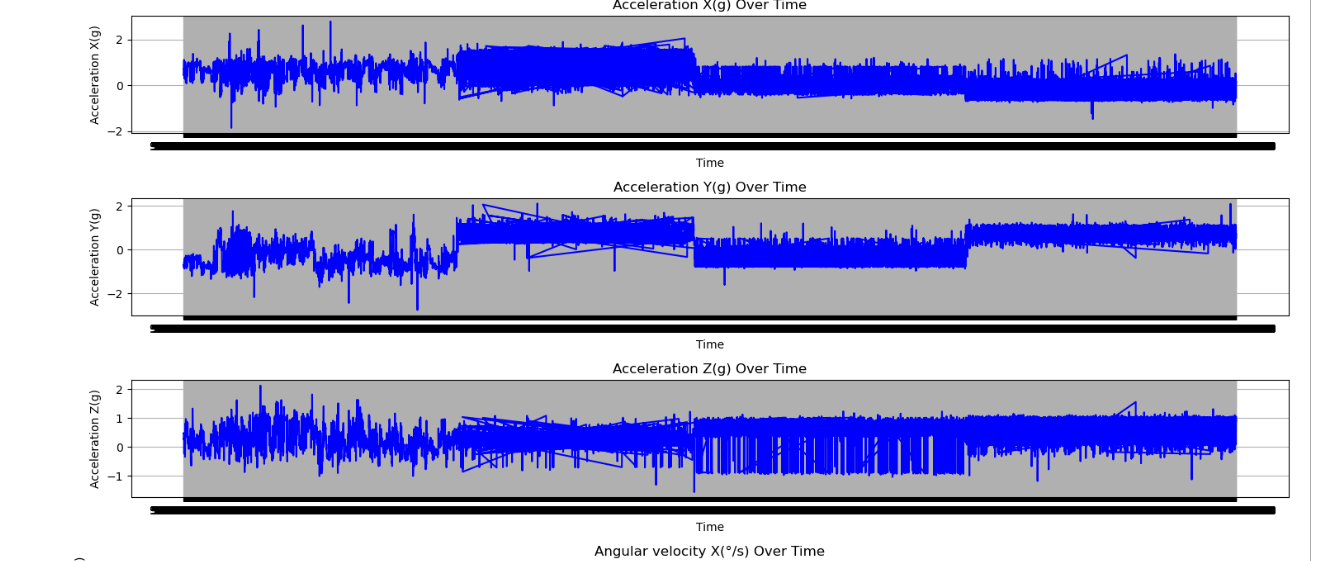}  
    \centering
    \includegraphics[width=\linewidth]{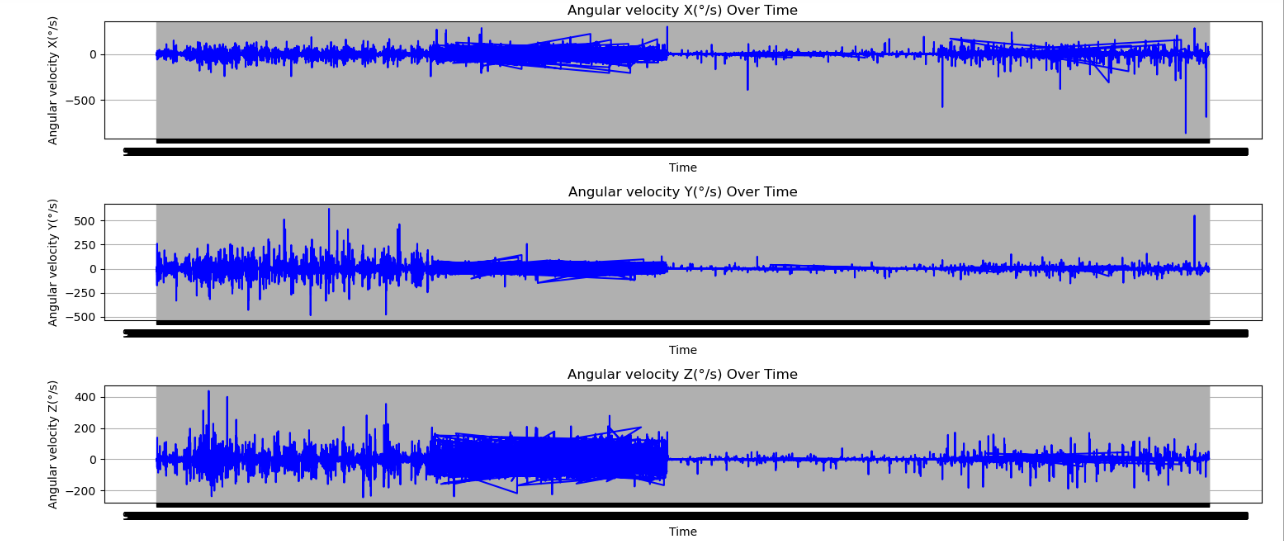}  
    \centering
    \includegraphics[width=\linewidth]{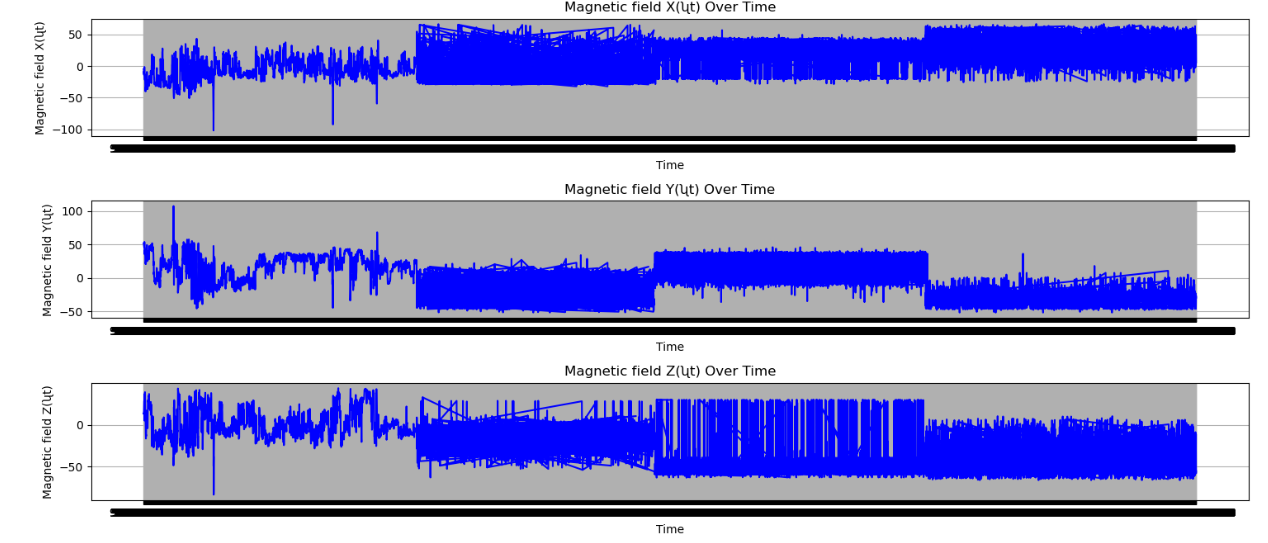}  
    \caption{Timestamp of dataset from the three different sensors (accelerometer, gyroscope and magnetometer) }
\end{figure}

The figures above provides a comprehensive view of how the sensor readings change over time. Each subplot represents the time series data for a specific sensor axis, enabling us to observe patterns, trends, and anomalies in the data collected during this study.

\begin{figure}[h!]
    \centering
    \includegraphics[width=\linewidth]{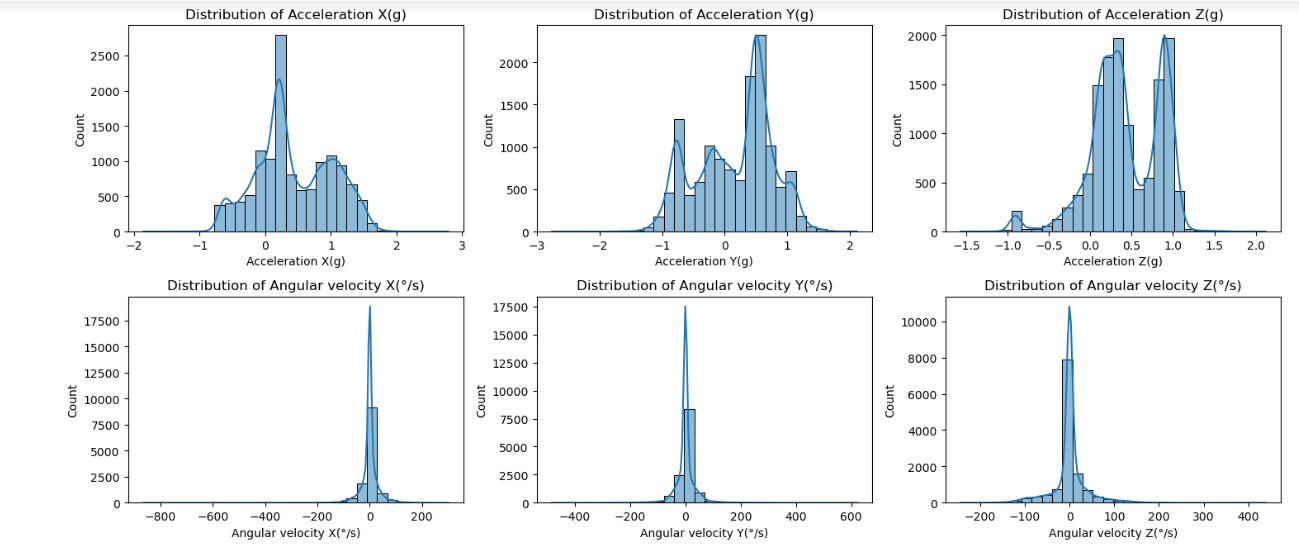}  
    \centering
    \includegraphics[width=\linewidth]{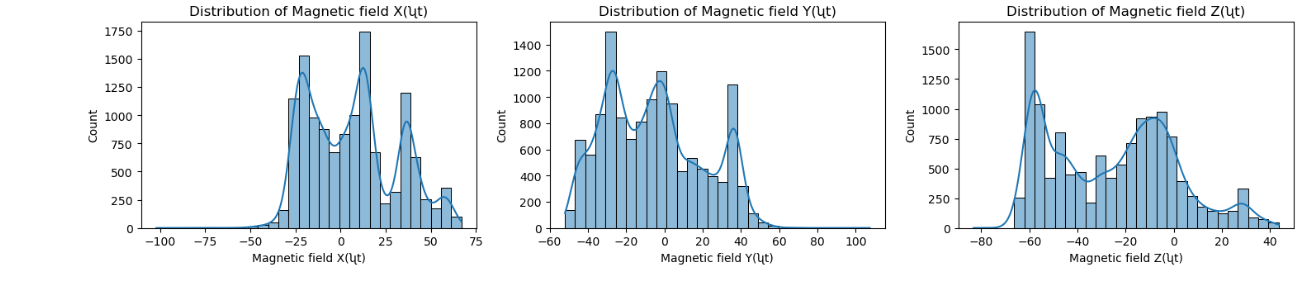}  
    \caption{Histogram for each sensor data collected}
\end{figure}
Figure 4 above shows the distribution of sensor readings from the accelerometer, gyroscope and magnetometer. This allows us to gain insights into the range and frequency of the values each sensor axis records. Each plot in the diagram indicates the data distribution for one of the sensor axes. The x-axis in each subplot represents the sensor reading values, while the y-axis shows the frequency of the values within our dataset. The Kernel Density Estimates, the line on top of each histogram plot, provide us with a smoothed version of the distribution if they are unimodal, bimodal or skewed.
Our plots clearly show that the angular velocity is an unimodal distribution, while our acceleration and magnetic field are both multimodal.
This unimodal distribution in angular velocity may indicate a consistent rotational movement during the monitored activities. 
The multimodal nature of the acceleration indicates that our subject experienced various types of movement and changes in velocity in the different activities. The magnetic field distribution could also result from exposure to varying magnetic environments or a frequent shift in the earth’s magnetic field during data collection at different positions.

\begin{figure}[h!]
    \centering
    \includegraphics[width=\linewidth]{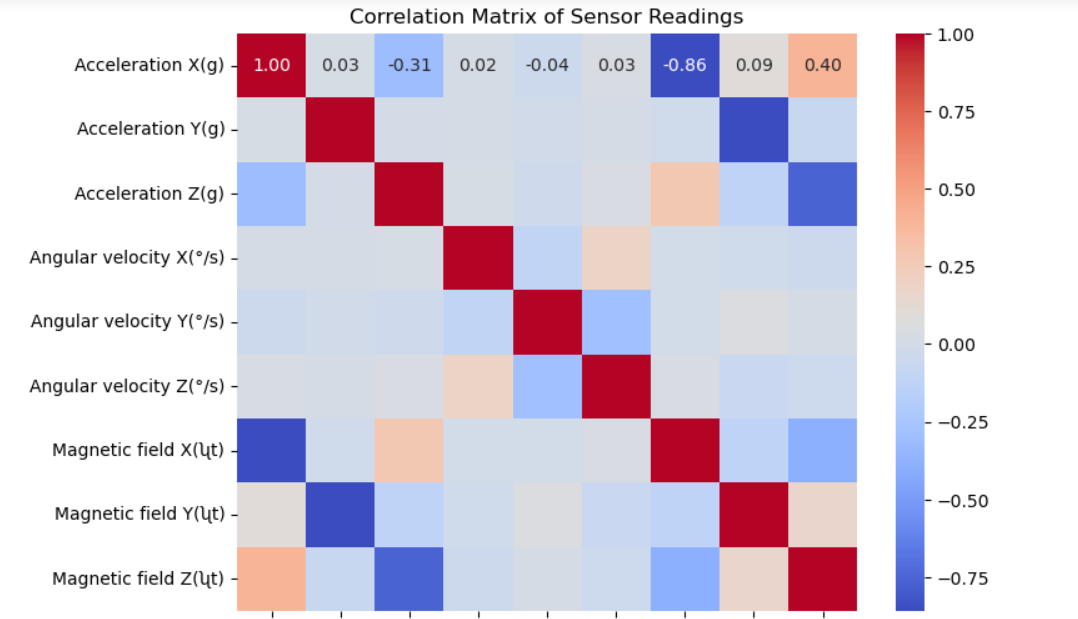}  
    \caption{Correlation matrix}
\end{figure}
The correlation matrix shows the correlation coefficients between pairs of variables from our sensor data, with a correlation coefficient ranging from -1(indicating a perfect negative correlation or inverse relation) to +1 (indicating an ideal correlation or direct relation). 
From the heat map in the diagram above, we can see that there is a very low correlation between pairs of variables. This indicates that there is low or no redundancy and that the variables in our data set could be potential complementary features and provide unique information for our activity recognition.
\begin{figure}[H]
    \centering
    \includegraphics[width=\linewidth]{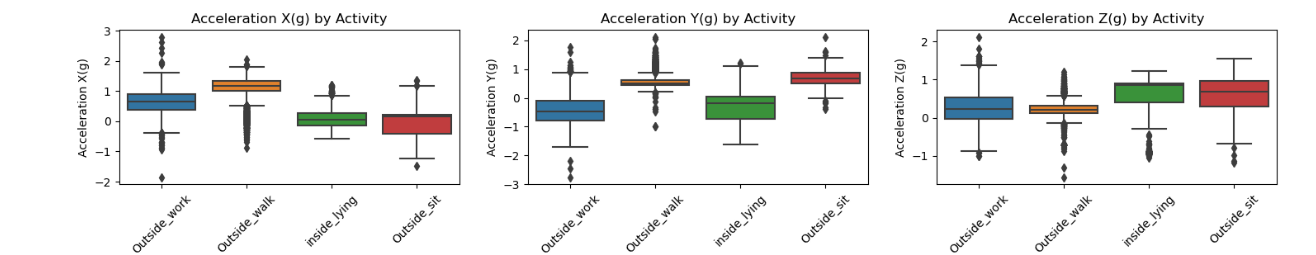}  
    \centering
    \includegraphics[width=\linewidth]{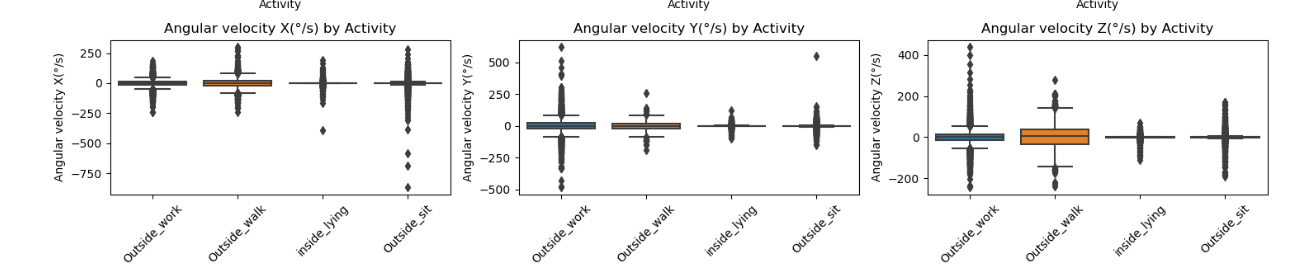}  
    \centering
    \includegraphics[width=\linewidth]{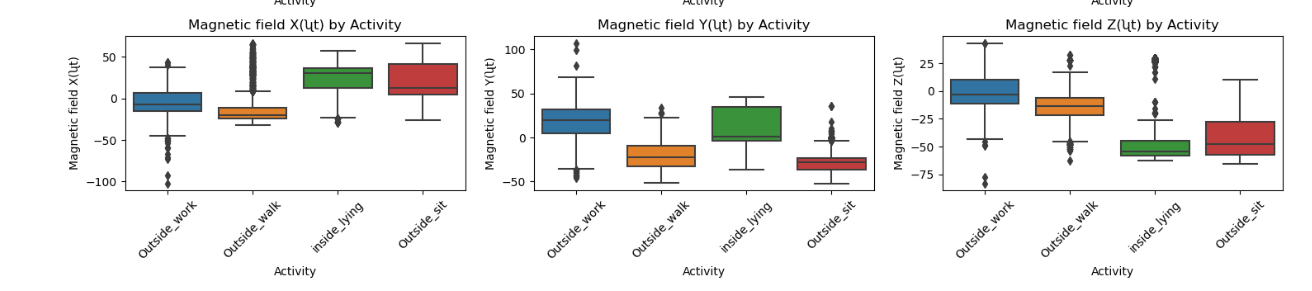}  
    \caption{Box Plot for each sensor data collected }
\end{figure}

\begin{figure}[H]
    \centering
    \includegraphics[width=\linewidth]{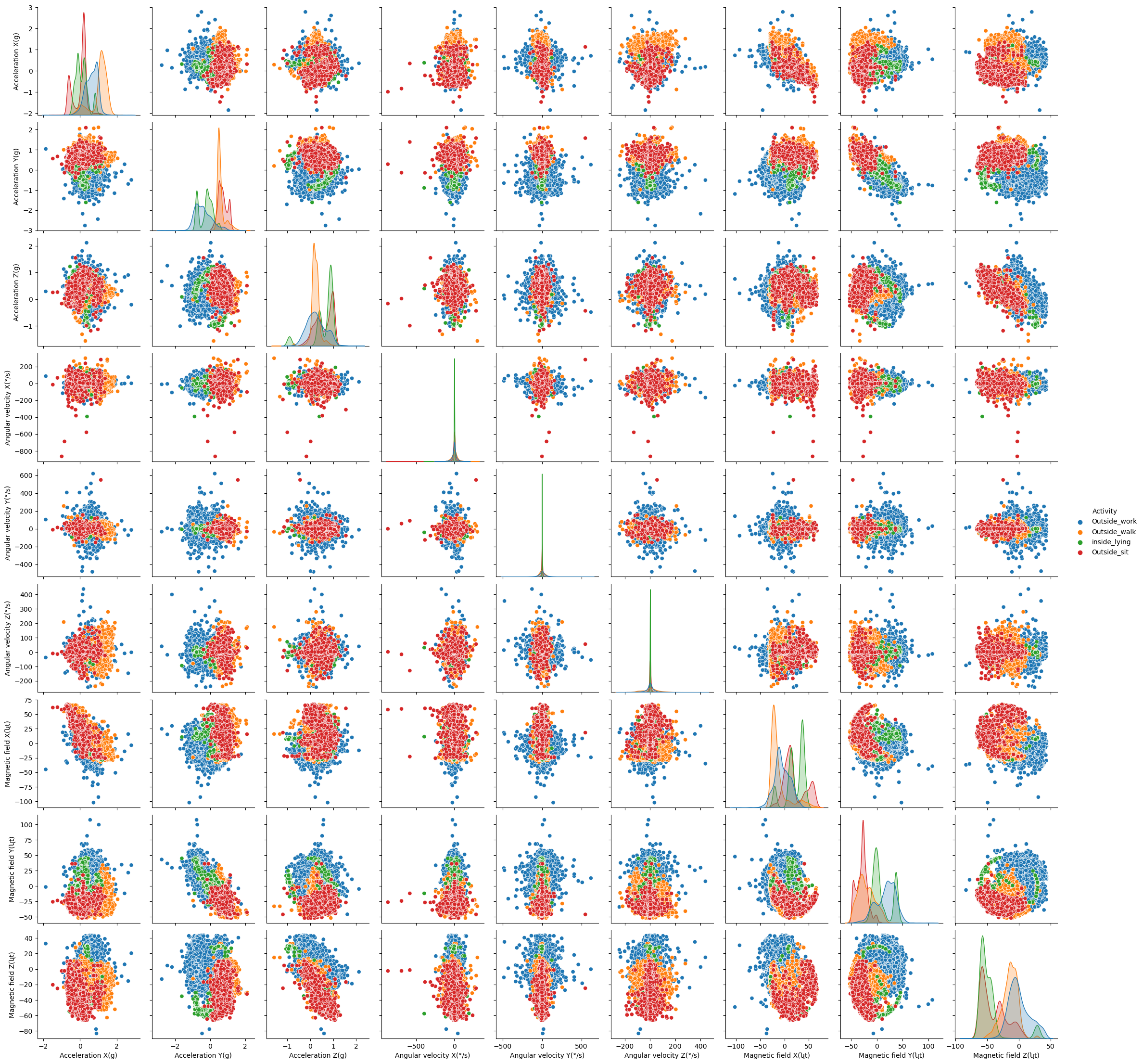}  
    \caption{Pair Plot for pair-wise relationship between sensors reading}
\end{figure}

A pairwise scatter plot of the relationship between sensor readings allows us to examine the relationship between different pairs of sensor features and how they vary across different activities. Along the positive diagonal, the histograms of each characteristic are displayed.
The pair plots above indicate that the four different activities form distinct clusters, which makes it easier for each activity to be identified for each sensor. This plot also gives us insight into the recorded data, and they appear to be outliers, anomalies, or rear movements, which are also present in the box plots. These outliers are seen as data points far from the main cluster in each pair plot.

\subsection{Individual Sensor Models}
The data was split into training and testing sets for each sensor type (accelerometer, gyroscope, magnetometer). The split ratio was 80\% for training and 20\% for model testing. Three different machine-learning algorithms were applied to the data from each sensor.
\begin{enumerate}
   
\item{Random Forest (RF)}
\item{Support Vector Machine (SVM)}
\item{Gradient Boosting (GB)}
\end{enumerate}
The machine learning models were chosen because they effectively handle complex patterns, non-linearities, and high-dimensional data. These models offer robustness, precision, and high accuracy, making them essential for interpreting the effect of data fusion on sensor data and for advanced applications across various domains.
Model Evaluation
The models' performance was evaluated using accuracy scores, confusion matrices, and the Root Mean-squared Error (RMSE).
\section*{Model 1: RANDOM FOREST (RF)}

\subsection*{Performance Metrics Summary}

\begin{table}[H]
\centering
\begin{tabular}{|l|l|l|l|l|l|}
\hline
\textbf{Sensor}      & \textbf{Accuracy} & \textbf{Precision} & \textbf{Recall} & \textbf{F1 Score} & \textbf{RMSE} \\ \hline
Accelerometer Data   & 0.9402            & 0.9405             & 0.9402          & 0.9402            & 0.1504        \\ \hline
Gyroscope Data       & 0.6188            & 0.6169             & 0.6188          & 0.6177            & 0.3535        \\ \hline
Magnetometer Data    & 0.9861            & 0.9862             & 0.9861          & 0.9861            & 0.0810        \\ \hline
\end{tabular}
\caption{Performance Metrics for Random Forest Model across different sensors}
\label{tab:rf_metrics}
\end{table}

\subsection*{Random Forest Confusion Matrix Breakdown}
\begin{longtable}{|l|l|c|c|c|c|c|}
\hline
\textbf{Sensor} & \textbf{Class} & \textbf{TP} & \textbf{FP} & \textbf{FN} & \textbf{TN} & \textbf{Accuracy} \\ \hline
Accelerometer & 0 & 616 & 50 & 33 & 1893 & \\ 
 & 1 & 624 & 48 & 40 & 1880 & \\ 
 & 2 & 606 & 36 & 51 & 1899 & \\ 
 & 3 & 591 & 21 & 31 & 1949 & 0.9402 \\ \hline
Gyroscope & 0 & 301 & 331 & 348 & 1612 & \\ 
 & 1 & 494 & 205 & 170 & 1723 & \\ 
 & 2 & 409 & 240 & 248 & 1695 & \\ 
 & 3 & 400 & 212 & 222 & 1758 & 0.6188 \\ \hline
Magnetometer & 0 & 641 & 10 & 8 & 1933 & \\ 
 & 1 & 651 & 18 & 13 & 1910 & \\ 
 & 2 & 645 & 4 & 12 & 1931 & \\ 
 & 3 & 619 & 4 & 3 & 1966 & 0.9861 \\ \hline
\end{longtable}

\section*{Model 2: Support Vector Machine (SVM)}

\subsection*{Performance Metrics Summary}

\begin{table}[h!]
\centering
\begin{tabular}{|l|l|l|l|l|l|}
\hline
\textbf{Sensor}      & \textbf{Accuracy} & \textbf{Precision} & \textbf{Recall} & \textbf{F1 Score} & \textbf{RMSE} \\ \hline
Accelerometer Data   & 0.9066            & 0.9110             & 0.9066          & 0.9059            & 0.1823        \\ \hline
Gyroscope Data       & 0.4336            & 0.5054             & 0.4336          & 0.3901            & 0.3899        \\ \hline
Magnetometer Data    & 0.9244            & 0.9239             & 0.9236          & 0.9235            & 0.1751        \\ \hline
\end{tabular}
\caption{Performance Metrics for Support Vector Machine Model across different sensors}
\label{tab:svm_metrics}
\end{table}


\section*{SVM Confusion Matrix Breakdown}

\begin{table}[H]
\centering
\begin{tabular}{|c|c|c|c|c|c|}
\hline
\textbf{Sensor and accuracy} & \textbf{Class} & \textbf{TP} & \textbf{FP} & \textbf{FN} & \textbf{TN} \\ \hline
\multirow{4}{*}{Accelerometer: 0.9066 } & 0 & 617 & 84  & 32  & 1859 \\ 
                               & 1 & 596 & 54  & 68  & 1874 \\
                               & 2 & 533 & 15  & 124 & 1920 \\
                               & 3 & 604 & 89  & 18  & 1881 \\ \hline
\multirow{4}{*}{Gyroscope: 0.4336}    & 0 & 39  & 171 & 610 & 1772 \\
                               & 1 & 318 & 101 & 346 & 1827 \\
                               & 2 & 165 & 66  & 492 & 1869 \\
                               & 3 & 602 & 1130 & 20  & 840  \\ \hline
\multirow{4}{*}{Magnetometer: 0.9236} & 0 & 610 & 59  & 39  & 1884 \\
                               & 1 & 578 & 80  & 86  & 1848 \\
                               & 2 & 602 & 25  & 55  & 1910 \\
                               & 3 & 604 & 34  & 18  & 1936 \\ \hline
\end{tabular}
\caption{Confusion Matrix Breakdown for SVM Model}
\end{table}

\section*{Model 3: GRADIENT BOOST (GB)}

\subsection*{Performance Metrics Summary}

\begin{table}[H]
\centering
\begin{tabular}{|l|l|l|l|l|l|}
\hline
\textbf{Sensor}      & \textbf{Accuracy} & \textbf{Precision} & \textbf{Recall} & \textbf{F1 Score} & \textbf{RMSE} \\ \hline
Accelerometer Data   & 0.9244            & 0.9248             & 0.9244          & 0.9242            & 0.1659        \\ \hline
Gyroscope Data       & 0.6200            & 0.6137             & 0.6200          & 0.6158            & 0.3498        \\ \hline
Magnetometer Data    & 0.9595            & 0.9596             & 0.9595          & 0.9595            & 0.1290        \\ \hline
\end{tabular}
\caption{Performance Metrics for Gradient Boost Model across different sensors}
\label{tab:gb_metrics}
\end{table}

\subsection*{Gradient Boost Confusion Matrix Breakdown}
\begin{longtable}{|l|l|c|c|c|c|c|}
\hline
\textbf{Sensor} & \textbf{Class} & \textbf{TP} & \textbf{FP} & \textbf{FN} & \textbf{TN} & \textbf{Accuracy} \\ \hline
Accelerometer & 0 & 611 & 62 & 38 & 1881 & \\ 
 & 1 & 608 & 53 & 56 & 1875 & \\ 
 & 2 & 583 & 34 & 47 & 1901 & \\ 
 & 3 & 594 & 47 & 28 & 1923 & 0.9244 \\ \hline
Gyroscope & 0 & 271 & 287 & 378 & 1656 & \\ 
 & 1 & 483 & 214 & 181 & 1714 & \\ 
 & 2 & 413 & 249 & 244 & 1686 & \\ 
 & 3 & 440 & 235 & 182 & 1735 & 0.6200 \\ \hline
Magnetometer & 0 & 626 & 33 & 23 & 1910 & \\ 
 & 1 & 617 & 49 & 47 & 1879 & \\ 
 & 2 & 628 & 13 & 29 & 1922 & \\ 
 & 3 & 616 & 10 & 6 & 1969 & 0.9595 \\ \hline
\end{longtable}
The first step in our evaluation process involved assessing the accuracy of models trained on individual sensor data (accelerometer, gyroscope, and magnetometer). This provides a baseline understanding of how each sensor contributes to activity recognition.\\
The Random Forest Model with Magnetometer data outperforms other models, as seen by its high scores in all metrics and a highly favourable confusion matrix.
The Support Vector Machine Model with the Gyroscope data performs the worst. This is also reflected in the low accuracy, precision, recall, F1 score, Highest RMSE value, and confusion matric values.
\begin{figure}[H]
    \centering
    \includegraphics[width=\linewidth]{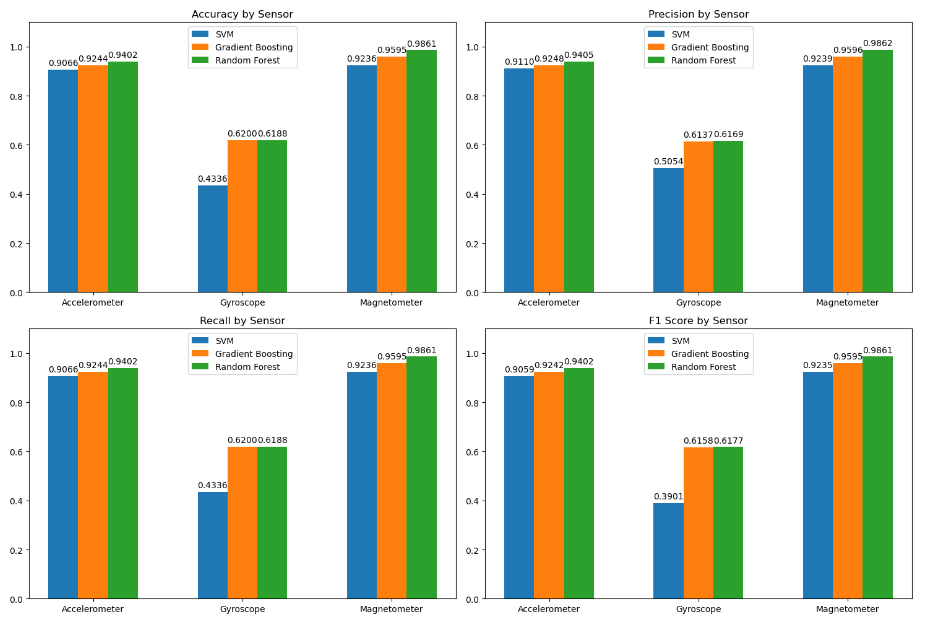}  

    \centering
    \includegraphics[width=\linewidth]{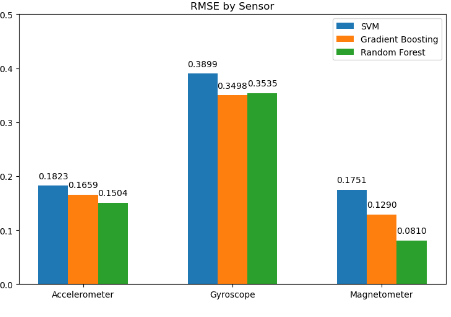}  
    \caption{Bar Plot of Accuracy, Precision, Recall, F1 Score, and RMSE scores.}
\end{figure}

\subsection{Feature Level  Fusion}Feature-level fusion involved combining the data from all three sensors (accelerometer, gyroscope, and magnetometer) into a single feature set. We used our three different models to analyse this combined sensor data to see its effects and accuracy.\\ The accelerometer, gyroscope, and magnetometer features were combined into a single feature set. The selected choice of models (Random Forest, support vector machine and the gradient boost models) was trained on the combined feature set. The performance was assessed using accuracy score, confusion matrices, and RMSE.\\
\section*{Model 1B: Random Forest  (RF) On Feature Fusion Data}

\textbf{Data Used:} Combined Data

\subsection*{Performance Metrics Summary}

\begin{table}[h!]
\centering
\begin{tabular}{|l|l|l|l|l|l|}
\hline
\textbf{Sensor}      & \textbf{Accuracy} & \textbf{Precision} & \textbf{Recall} & \textbf{F1 Score} & \textbf{RMSE} \\ \hline
Accelerometer Data   & 0.9402            & 0.9405             & 0.9402          & 0.9402            & 0.1504        \\ \hline
Gyroscope Data       & 0.6188            & 0.6169             & 0.6188          & 0.6177            & 0.3535        \\ \hline
Magnetometer Data    & 0.9861            & 0.9862             & 0.9861          & 0.9861            & 0.0810        \\ \hline
Combined Data        & 0.9811            & 0.9811             & 0.9811          & 0.9811            & 0.0884        \\ \hline
\end{tabular}
\caption{Performance Metrics for Random Forest Model across different sensors and combined data}
\label{tab:rf_metrics_combined}
\end{table}

\begin{figure}[H]
    \centering
    \includegraphics[width=\linewidth]{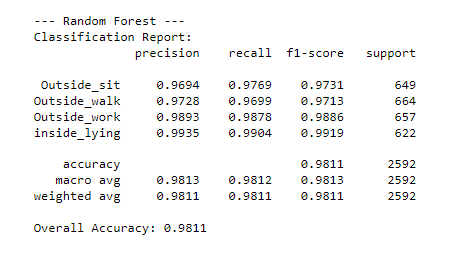}  
\end{figure}

\begin{table}[h!]
\centering
\resizebox{\textwidth}{!}{ 
\begin{tabular}{|c|c|c|c|c|c|c|c|c|c|}
\hline
\textbf{MODEL} & \textbf{TP} & \textbf{TN} & \textbf{FP} & \textbf{FN} & \textbf{PRECISION} & \textbf{RECALL} & \textbf{F1-SCORE} & \textbf{RMSE} & \textbf{ACCURACY} \\ \hline
SVM            & 2510        & 7694        & 82         & 82         & 0.9687            & 0.9684          & 0.9684           & 0.1087        & 0.9684          \\ \hline
G/BOOST        & 2525        & 7709        & 67         & 67         & 0.9742            & 0.9742          & 0.9742           & 0.0961        & 0.9742          \\ \hline
RF             & 2543        & 7727        & 49         & 49         & 0.9811            & 0.9811          & 0.9811           & 0.0884        & 0.9811          \\ \hline
\end{tabular}
}
\caption{Summary of Models Result on Feature Fusion and Confusion Matrix Breakdown}
\end{table}

\section*{Model 2B: Support Vector Machine (SVM) On Feature Fusion Data}

\textbf{Data Used:} Combined Data

\subsection*{Performance Metrics Summary}

\begin{table}[h!]
\centering
\begin{tabular}{|l|l|l|l|l|l|}
\hline
\textbf{Sensor}      & \textbf{Accuracy} & \textbf{Precision} & \textbf{Recall} & \textbf{F1 Score} & \textbf{RMSE} \\ \hline
Accelerometer Data   & 0.9066            & 0.9110             & 0.9066          & 0.9059            & 0.1823        \\ \hline
Gyroscope Data       & 0.4336            & 0.5054             & 0.4336          & 0.3901            & 0.3899        \\ \hline
Magnetometer Data    & 0.9244            & 0.9239             & 0.9236          & 0.9235            & 0.1751        \\ \hline
Combined Data        & 0.9684            & 0.9687             & 0.9684          & 0.9684            & 0.1087        \\ \hline
\end{tabular}
\caption{Performance Metrics for Support Vector Machine Model across different sensors and combined data}
\label{tab:svm_metrics_combined}
\end{table}

\begin{figure}[H]
    \centering
    \includegraphics[width=\linewidth]{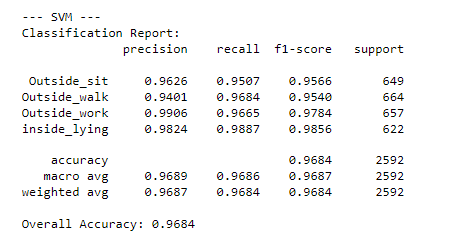}  
\end{figure}

\section*{Model 3B: Gradient Boost  (GB) On Feature Fusion Data}

\textbf{Data Used:} Combined Data

\subsection*{Performance Metrics Summary}

\begin{table}[h!]
\centering
\begin{tabular}{|l|l|l|l|l|l|}
\hline
\textbf{Sensor}      & \textbf{Accuracy} & \textbf{Precision} & \textbf{Recall} & \textbf{F1 Score} & \textbf{RMSE} \\ \hline
Accelerometer Data   & 0.9244            & 0.9248             & 0.9244          & 0.9242            & 0.1659        \\ \hline
Gyroscope Data       & 0.6200            & 0.6137             & 0.6200          & 0.6158            & 0.3498        \\ \hline
Magnetometer Data    & 0.9595            & 0.9596             & 0.9595          & 0.9595            & 0.1290        \\ \hline
Combined Data        & 0.9742            & 0.9742             & 0.9742          & 0.9742            & 0.0961        \\ \hline
\end{tabular}
\caption{Performance Metrics for Gradient Boost Model across different sensors and combined data}
\label{tab:gb_metrics_combined}
\end{table}

\begin{figure}[H]
    \centering
    \includegraphics[width=\linewidth]{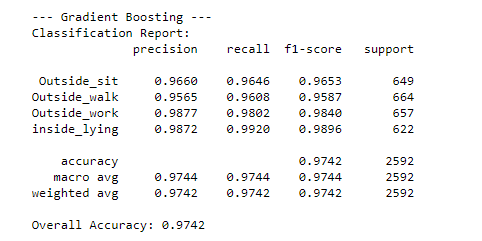}  
\end{figure}

The model's accuracy using combined sensor data indicates a high level of accuracy for all models. This underscores the success of feature-level fusion, instilling confidence in its effectiveness. 
From the table above, we can easily see a very high percentage of precision and recall values across the three models. The precision, indicating a high level of optimistic prediction, suggests that combining the diverse information provided by different sensors significantly enhances the model's ability to classify activities accurately.
\begin{figure}[H]
    \centering
    \includegraphics[width=\linewidth]{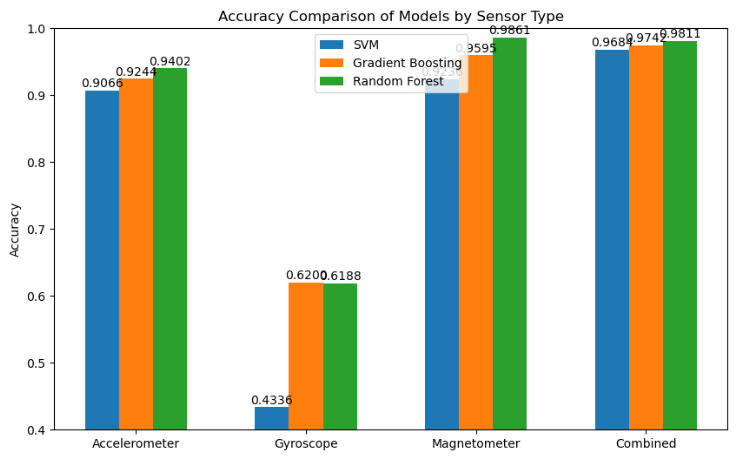}  
    \caption{ACCURACY COMPARISON OF MODELS BY SENSOR TYPE vs. COMBINED}
\end{figure}

The plot in Table above visually demonstrates that using combined data from all sensors significantly enhances accuracy across all models, particularly highlighting the superiority of the random forest when utilizing fused sensor data. 
The graph also highlights the advantages and power of data fusion. Data Fusion does not subtract from a model's accuracy or average out. Still, it adds more information not captured by other sensors, which can be seen in the gyroscope's accuracy from the individual sensor accuracy compared to the combined accuracy of any of the three models used.
Although it is seen that the magnetometer accuracy seems slightly higher than any of the combined models, a closer look at the confusion matrix of the random forest model used for the magnetometer and the random forest used for the combined (feature fusion gives a deeper insight on the benefits of data fusion.\\

\subsubsection{Decision Level Fusion:}Individual models were trained for each sensor as previously described. Majority voting was applied to combine the predictions from the individual models. The decision-level fusion approach was evaluated.\\
Decision-level fusion was implemented using the 'majority voting 'technique, which combines the predictions from separate models trained on individual sensors by selecting the most commonly predicted class.
Accuracy using Decision Fusion (Majority Voting): 97.80\%.
\textbf{Analysis} \\

\textbf{Table 9: Effect of Data Fusion on Three Models Used for Our Data Set on Activity Recognition.}

\begin{table}[h!]
\centering
\resizebox{\textwidth}{!}{%
\begin{tabular}{|l|l|l|l|l|l|}
\hline
\textbf{Model} & \textbf{Accelerometer} & \textbf{Gyroscope} & \textbf{Magnetometer} & \textbf{Feature Fusion} & \textbf{Decision-level Fusion (VOTING)} \\ \hline
SVM             & 0.9066 RMSE: 0.1823    & 0.4336 RMSE: 0.3899 & 0.9236 RMSE: 0.1751    & 0.9684 RMSE: 0.1087    & 0.9780 RMSE: 0.0910 \\ \hline
Gradient Boost  & 0.9244 RMSE: 0.1659    & 0.6200 RMSE: 0.3498 & 0.9595 RMSE: 0.1240    & 0.9742 RMSE: 0.0961    &  \\ \hline
Random Forest   & 0.9402 RMSE: 0.1504    & 0.6188 RMSE: 0.3535 & 0.9861 RMSE: 0.0810    & 0.9811 RMSE: 0.0884    &  \\ \hline
\end{tabular}%
}
\end{table}

The accuracy achieved through decision-level fusion is lower than that obtained via feature-level fusion but still higher than the accuracy of individual sensor models (except for the magnetometer with the Random Forest model). \\

\subsubsection{Kalman Filter Fusion:}The Kalman Filter was applied to fuse measurements from the accelerometer, gyroscope, and magnetometer. The Kalman Filter method is a well-known fusion method commonly used in robotics and autonomous systems; hence, it's worth looking at its performance using our dataset from our sensors. Two commonly used measurement fusion methods exist for Kalman-filter-based multisensory data fusion. The first (Method I) merges the multisensor data through the observation vector of the Kalman filter.
The second (Method II) combines the multisensor data based on a minimum-mean-square-error criterion.
According to \cite{gan2001}, both measurement fusion methods are functionally equivalent if the data fusion sensors with different independent noise characteristics have identical measurement matrices.
Implementing the Kalman filter for this project involves merging the sensor data through the observation vector. 

\cite{sasiadek2000} used an Extended Kalman Filter (EKF) to estimate the position of the mobile vehicle. In their work, the innovation and covariance of the innovation process are monitored using the Adaptive Fuzzy Logic System (AFLS) to prevent the filter from diverging, resulting in an adaptation gain of EKF.\\ Several feature engineering techniques could be  implemented to improve model performance such as\\
•	Statistical Features: Calculated each sensor's mean, variance, and standard deviation.\\
•	Time-Domain Features: Extracted features related to the time domain, and the\\
•	Windowing: Applied sliding windows to segment the data and capture temporal patterns. But these are not captured in this work.\\
Kalman Filter fusion was applied to combine sensor measurements over time, producing a filtered dataset that was used to train a Random Forest model.
Accuracy using Kalman Filtered Data: 95.25\%.
\begin{figure}[H]
    \centering
    \includegraphics[width=\linewidth]{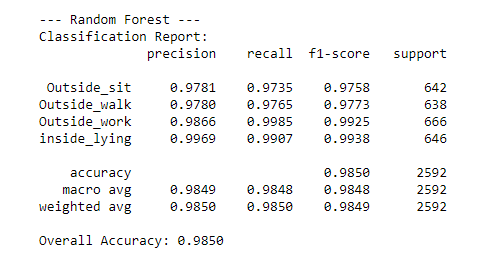}  
    \caption{Kalman Filter Confusion Matrix breakdown}
\end{figure}\
Kalman Filter fusion significantly improved accuracy compared to decision fusion but was slightly less effective than feature-level fusion with Random Fores. However, Kalman Filter fusion excels in giving a smooth estimate of the sensor data over time, which could benefit applications requiring continuous tracking like health care monitoring, sports analysis and navigations. The lower accuracy of the Kalman filter in relation to the feature fusion could be attributed to the fact that the latter captures more complex interactions between features, which could explain its higher accuracy.
\section{Evaluation of Results} This chapter evaluates the activity recognition models developed using various data fusion techniques. Each model's performance metrics, including accuracy, precision, recall, and F1-score, were calculated to assess its effectiveness. This evaluation is crucial to determining the impact of different sensor data and fusion methodologies on the accuracy of activity classification.

\subsection{Analysis of Results And Findings}The results indicate that the magnetometer provided the highest accuracy across all models, with the Random Forest model performing exceptionally well. In contrast, the gyroscope data resulted in lower accuracy, suggesting that it is less effective as a standalone sensor for activity recognition in this context.

\textbf{Overall Improved Robustness and Accuracy} \\
- Magnetometer with RF: On our class of four, (sitting, walking, working and lying), the magnetometer has a sum of 36 as False Negative, as shown below
\begin{figure}[H]
    \centering
    \includegraphics[width=\linewidth]{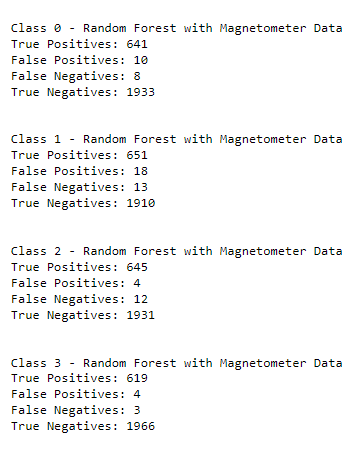}  
\end{figure}
While the individual sensor (magnetometer) seems to perform well under ideal conditions, combining data from multiple sensors (accelerometer, gyroscope, and magnetometer) leads to an overall more robust and reliable system in varying conditions. This fusion process has provided a buffer against the weakness of any one sensor, leading to consistent performance across different environments. 
The disparity among classes in the magnetometer result is evident from the figure above, as seen in the performance of Class 1 and Class 3 (with 18 FP, 13 FN for Class 1 and 4 FP, 3FN for Class 3). This contrasts with the result from the Data fusion, which typically leads to a more balanced performance across all classes, as shown below.
\begin{figure}[H]
    \centering
    \includegraphics[width=\linewidth]{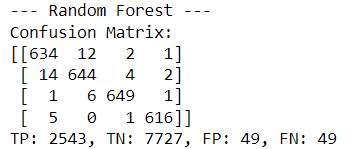}  
\end{figure}

\begin{table}[H]
\caption{Sensor-level accuracy and combined sensor data accuracy for various models.}
\begin{tabularx}{\textwidth}{|l|X|X|X|}
\hline
\textbf{Model} & \textbf{Sensor-level Accuracy} & \textbf{Feature-level Fusion} & \textbf{Decision-level Fusion} \\ \hline
SVM & \begin{tabular}[c]{@{}l@{}}ACC: 0.9066\\ GYR: 0.4336\\ MAG: 0.9236\end{tabular} & 0.9684 & 0.9780 \\ \hline
GBOOST & \begin{tabular}[c]{@{}l@{}}ACC: 0.9244\\ GYR: 0.6200\\ MAG: 0.9595\end{tabular} & 0.9742 &  \\ \hline
R/FOREST & \begin{tabular}[c]{@{}l@{}}ACC: 0.9402\\ GYR: 0.6188\\ MAG: 0.9861\end{tabular} & 0.9811 &  \\ \hline
\end{tabularx}
\end{table}

\textbf{Improvement through Data Fusion}\\ The accuracy increased slightly when combining data from all three sensors. This suggests that integrating different types of information provides a more complete picture and allows more precision in activity recognition.
The table above shows that the magnetometer alone was very effective, which shows the modest data fusion approach, as each sensor contributes unique and valuable information.
The evaluation results also demonstrate the importance of data fusion in improving model accuracy, providing a more complete picture that is very important in the aspect of digital twins. While individual sensors can provide good accuracy, as seen from the single magnetometer accuracy, the data fusion approach still offers an edge in our experiment, capturing more nuanced information and leading to better performance.
Using multiple sensors and fusing their data in real-world scenarios could significantly enhance the system's reliability. Ensuring consistency and accurate classification is more important than maximizing accuracy in ideal conditions. Our project work shows that data fusion ensures that your model is more likely to make the right decision even when some sensors are compromised, leading to a more reliable overall system.\\
\textbf{Highest single sensor Accuracy:} Magnetometer with Random Forest performed the best among single sensor models (98.61\%). \\

\textbf{Feature Fusion Accuracy:} 98.11\%. \\

\textbf{Decision Fusion:} 97.80\%, demonstrating robustness but slightly less effective than feature-level fusion. \\

\textbf{Kalman Filter Fusion:} 95.25\%, striking a balance between accuracy and temporal consistency, providing reassurance about its performance. \\
The evaluation of results indicates that combining sensor data through feature-level fusion significantly enhances the performance of activity recognition models. While decision-level and Kalman Filter fusion also improve accuracy compared to individual sensor models, feature-level fusion with Random Forest proved to be the most effective approach in this study. \\

The findings underscore the significance of integrating diverse sensor data to achieve higher classification accuracy in activity recognition tasks. \\
\begin{figure}[H]
    \centering
    \includegraphics[width=\linewidth]{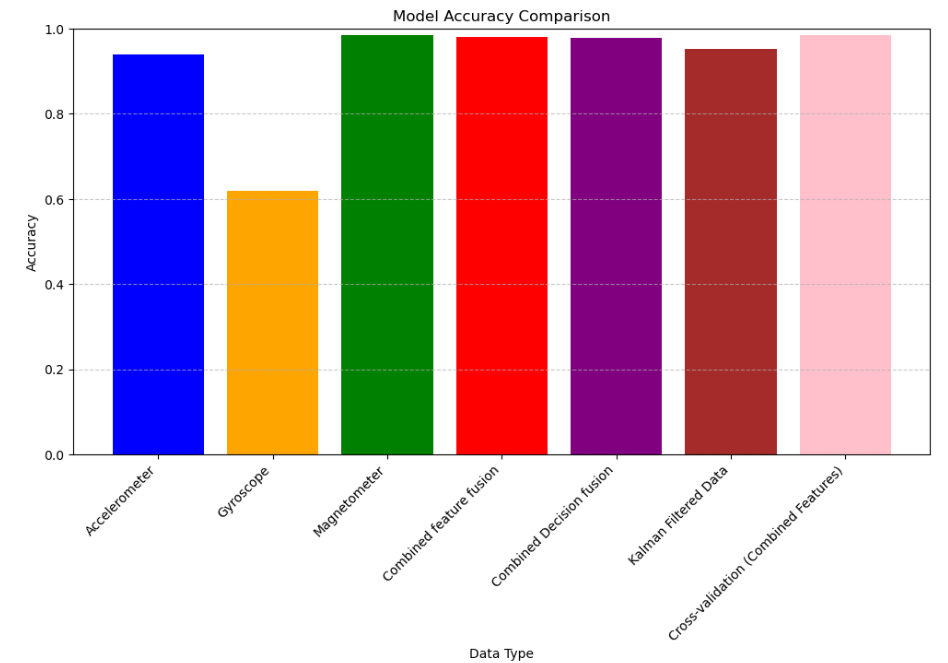}  
    \caption{Histogram of the accuracy of results on Sensors Data}
\end{figure}\
I would also like to point out that this paper emphasizes the importance of data fusion for model enhancement. So, the models used and their parameters were not hyper-tuned for better performance. Their default state and kernel were used across the data sets and fusion process, and at this stage, all accuracy and performance were estimated.

\subsection{Comparison Of Results}
The algorithms used for the paper were implemented to evaluate the efficacy of data fusion using a secondary dataset Published on 20 November 2020 and generated by contributors Ivan Pires and Nuno M. Garcia, (\cite{pires2020}). This was done for comparison purposes.
This dataset presents the data related to walking, running, standing, walking upstairs, and walking downstairs, captured with accelerometer, gyroscope, and magnetometer sensors available in off-the-shelf mobile devices. The dataset can be accessed at:\\ \href{https://data.mendeley.com/datasets/xknhpz5t96/2}{https://data.mendeley.com/datasets/xknhpz5t96/2}.
Data synchronization was carried out on the secondary dataset with its timestamp on the sensor's data and their activities to allow the fusion process.
\begin{figure}[H]
    \centering
    \includegraphics[width=\linewidth]{"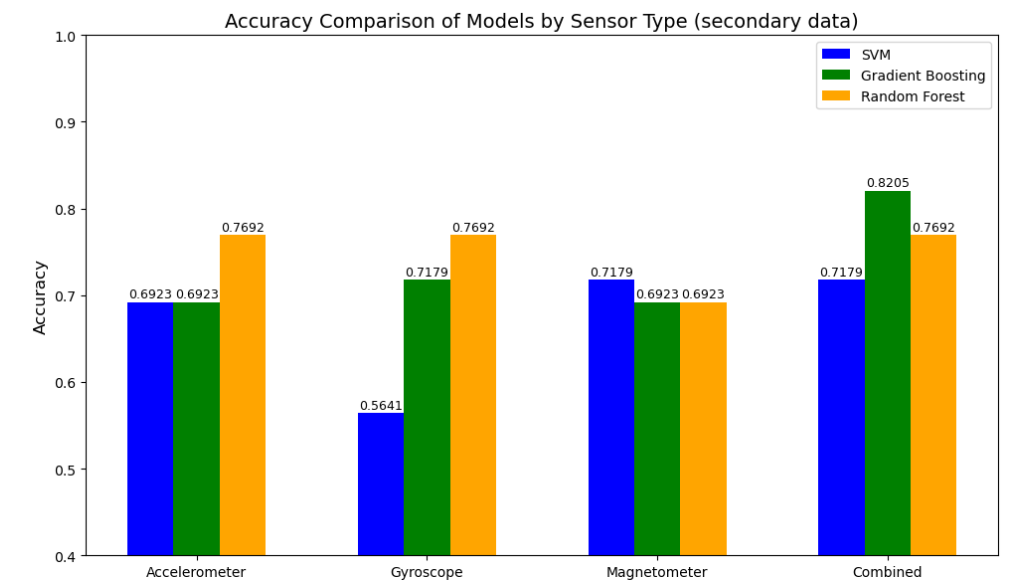"}  
\end{figure}\
\begin{figure}[H]
    \centering
    \includegraphics[width=\linewidth]{"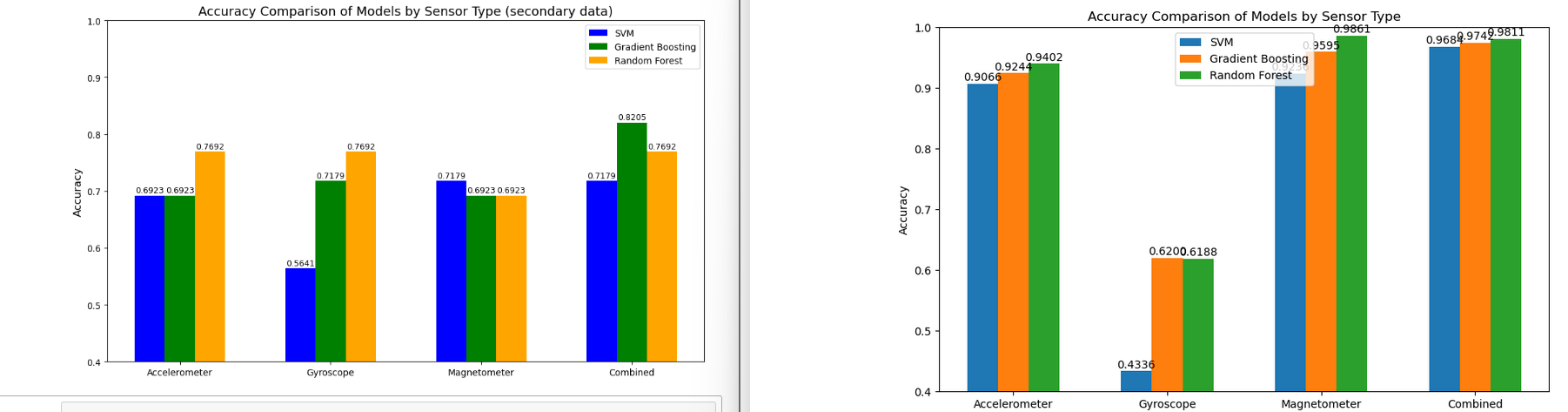"}  
    \caption{Secondary Vs Primary Dataset for the accuracy of results on algorithms used for Sensors Data}
\end{figure}\
The dataset from Ivan Pires and Nuno M. Garcia shows results similar to our primary dataset, which has the highest accuracy on the fused dataset. This further iterated the importance of data fusion and cannot be over-emphasized.\\
\begin{figure}[H]
    \centering
    \begin{subfigure}[t]{0.45\textwidth}
        \centering
        \includegraphics[width=\linewidth]{"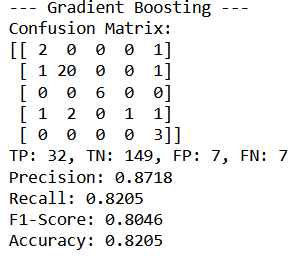"}  
        \caption{Secondary confusion matrix for fused data showing stable performance}  
        \label{fig:secondary_dataset_confusion}
    \end{subfigure}
    \hfill
    \begin{subfigure}[t]{0.45\textwidth}
        \centering
        \includegraphics[width=\linewidth]{"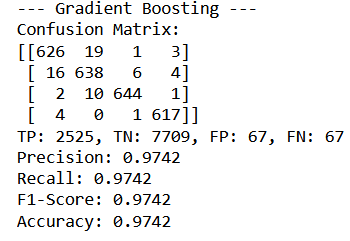"}  
        \caption{Primary confusion matrix for fused data showing stable performance}
        \label{fig:secondary_vs_primary_confusion}
    \end{subfigure}
    \caption{Comparison of secondary dataset confusion matrix and secondary vs primary fused data confusion matrix.}
    \label{fig:comparison_figures}
\end{figure}
This fusion process confusion matrix from the secondary dataset stands as a validation of the buffering effect of data fusion against the
weakness of any one sensor, leading to consistent performance across different environments. This is noticed in the stability of the FP and FN from both datasets( primary and secondary).

\subsection{Data Privacy And Ethics}While some existing studies take data privacy and security into account, it's important to note that in Machine learning, large data sets are crucial to ensure learning quality and fusion accuracy. However, the use of original data in machine learning could potentially lead to sensitive information leakage. This risk is particularly acute in Internet-related areas such as intrusion detection, attack analysis, and location tracking. It's our collective responsibility to manage this risk and ensure the privacy and security of data providers.\\
This study aimed to enhance the performance of activity recognition models by harnessing the potential of data fusion techniques, specifically in the context of a home environment. By combining sensor data from accelerometers, gyroscopes, and magnetometers, we sought to significantly improve the accuracy and robustness of machine learning models that classify activities such as walking, working, sitting, and lying.
The evaluation demonstrated that data fusion significantly enhances the model's ability to recognize activities. Key findings include:

\textbf{Individual Sensor Performance:} The accuracy of models trained on data from individual sensors varied significantly. While the magnetometer data provided the highest accuracy among single sensors, the gyroscope data performed the worst. This variability highlights the limitations of relying on a single type of sensor for activity recognition. \\

\textbf{Feature-Level Fusion:} Combining sensor data into a single feature set significantly improved classification accuracy. The Random Forest model trained on the combined data achieved an accuracy of 0.9811 (98.11\%), outperforming models trained on individual sensor data of accelerometers and gyroscopes. This underscores the effectiveness of feature-level fusion in enhancing model performance. \\

\textbf{Decision-Level Fusion:} The decision fusion approach aggregated predictions from separate sensor models using majority voting and yielded a high accuracy of 0.9780 (97.80\%). However, it was slightly less effective than feature-level fusion, suggesting that integrating sensor data at the feature level provides more complementary information to the model. \\

\textbf{Kalman Filter Fusion:} The application of the Kalman Filter to fuse sensor measurements further demonstrated the benefits of data fusion. With an overall accuracy of 0.9525 (95.25\%) and strong performance across all activity classes, this method proved to be a robust and reliable approach for activity recognition, particularly in environments with potential sensor noise, providing reassurance about its performance. \\

Overall, this study confirms that data fusion techniques, especially feature-level and Kalman Filter fusion, are critical for improving the accuracy and reliability of activity recognition models. Integrating multiple sensor modalities allows the models to capture a broader and more nuanced understanding of activities, leading to more accurate predictions.\\
\section{Conclusion And Future Works}
In conclusion, this paper has demonstrated the significant potential of data fusion techniques in improving activity recognition models. By effectively combining sensor data, we can build more accurate and robust models better suited for real-world applications in smart homes using wearable technologies and other IoTs. This has the potential to significantly impact the field of human activity monitoring, sparking excitement and hope for the future. \\

While more advanced models are available that may be well suited for fusion, our research has shown the adaptability of data fusion using machine models such as the Support Vector Machine, the Gradient Boost, and the Random Forest. When integrated with data fusion techniques like the Kalman Filter, these models have shown considerable promise in enhancing accuracy and robustness. The fusion techniques used are one of many available types that could be explored in future work, providing a sense of confidence in the adaptability of our research. \\
While this study has shown promising results, there are several exciting avenues for future research and improvement. Exploring Additional Sensors, Advanced Data Fusion Techniques, Real-Time Activity Recognition, Transfer Learning and Personalization, and Handling Imbalanced Data are all promising areas that could significantly advance the field of activity recognition. \\
Future work could explore the integration of additional sensors, such as barometers or light sensors, cameras especially from different devices unlike the WitMotion devices which capture different parameter data on a go. This approach will further enrich the data and potentially capture more complex activities or contextual information. \\
There is scope to investigate more advanced data fusion techniques, such as deep learning-based fusion methods, which can automatically learn the most effective way to combine sensor data. Techniques like Convolutional Neural Networks (CNNs) and Recurrent Neural Networks (RNNs) could be explored. Future studies could also explore applying Kalman Filter fusion to other types of sensor data or activities to access more generalized information about the model. \\

Implementing the models in real-time systems could be an exciting next step. This would involve optimizing the models for faster inference and potentially integrating them into wearable devices or smart home systems. \\
Another exciting area for future research is the application of transfer learning to adapt the models to different individuals or environments. This could involve fine-tuning the models based on user-specific data, allowing for personalized activity recognition that accounts for individual differences in movement patterns. \\

While this study focused on four balanced activity classes, real-world data often involves imbalanced classes. Future work could explore techniques for handling imbalanced data, such as synthetic data generation, cost-sensitive learning, or advanced resampling methods. This presents a significant challenge that requires innovative solutions. \\
Also, conducting longitudinal studies to evaluate these models' long-term performance and usability in real-world settings would provide valuable insights, especially regarding the Kalman Filter Fusion model. Understanding how these models perform over time and in different conditions is crucial for further refinements and ensuring their practical applicability. \\


\section*{References}

\noindent
[1] M. Attaran and B.G. Celik. “Digital Twin: Benefits, use cases, challenges, and opportunities”. In: \textit{Decision Analytics Journal} 6 (2023), p. 100165. doi: \url{https://doi.org/10.1016/j.dajour.2023.100165}.

[2] Mahmoud A Bakr and Seungchul Lee. “Distributed Multisensor Data Fusion under Unknown Correlation and Data Inconsistency”. In: \textit{Sensors} 17.11 (2017), p. 2472. doi: \url{10.3390/s17112472}.

[3] E. Blasch et al. “High Level Information Fusion developments, issues, and grand challenges: Fusion 2010 panel discussion”. In: \textit{2010 13th International Conference on Information Fusion}. 2010, pp. 1–8. doi: \url{10.1109/ICIF.2010.5712116}.

[4] F. Castanedo. “A Review of Data Fusion Techniques”. In: \textit{The Scientific World Journal} 2013 (2013), pp. 1–19. doi: \url{https://doi.org/10.1155/2013/704504}.

[5] D. De Kerckhove. “The personal digital twin, ethical considerations”. In: \textit{Philosophical Transactions of the Royal Society A: Mathematical, Physical and Engineering Sciences} 379.2207 (2021), p. 20200367. doi: \url{https://doi.org/10.1098/rsta.2020.0367}.

[6] Q. Gan and C.J. Harris. “Comparison of two measurement fusion methods for Kalman-filter-based multisensor data fusion”. In: \textit{IEEE Transactions on Aerospace and Electronic Systems} 37.1 (2001), pp. 273–279. doi: \url{https://doi.org/10.1109/7.913685}.

[7] B. He, X. Cao, and Y. Hua. “Data fusion-based sustainable digital twin system of intelligent detection robotics”. In: \textit{Journal of Cleaner Production} 280 (2021), p. 124181. doi: \url{10.1016/j.jclepro.2020.124181}.

[8] D.-Y. et al. Jeong. “Digital Twin: Technology Evolution Stages and Implementation Layers With Technology Elements”. In: \textit{IEEE Access} 10 (2022), pp. 52609–52620. doi: \url{https://doi.org/10.1109/ACCESS.2022.3174220}.

[9] D. et al. Jones. “Characterising the Digital Twin: A systematic literature review”. In: \textit{CIRP Journal of Manufacturing Science and Technology} 29 (2020), pp. 36–52. doi: \url{https://doi.org/10.1016/j.cirpj.2020.02.002}.

[10] Bahador Khaleghi et al. “Multisensor data fusion: A review of the state-of-the-art”. In: \textit{Information Fusion} 14.1 (2013), pp. 28–44. doi: \url{10.1016/j.inffus.2011.08.001}.

[11] W. et al. Kritzinger. “Digital Twin in manufacturing: A categorical literature review and classification”. In: \textit{IFAC-PapersOnLine} 51.11 (2018), pp. 1016–1022. doi: \url{https://doi.org/10.1016/j.ifacol.2018.08.474}.

[12] Martin E. Liggins, James Llinas, and David L. Hall, eds. \textit{Handbook of Multisensor Data Fusion: Theory and Practice}. 2nd ed. The Electrical Engineering and Applied Signal Processing Series. Boca Raton, FL: CRC Press, 2009, p. 849. ISBN: 978-1-4200-5308-1.

[13] A. Macías et al. “Data fabric and digital twins: An integrated approach for data fusion design and evaluation of pervasive systems”. In: \textit{Information Fusion} 103 (2024), p. 102139. doi: \url{https://doi.org/10.1016/j.inffus.2023.102139}.

[14] T. Meng et al. “A survey on machine learning for data fusion”. In: \textit{Information Fusion} 57 (2020), pp. 115–129. doi: \url{10.1016/j.inffus.2019.12.001}.

[15] H. Mengyan et al. “Current status of digital twin architecture and application in nuclear energy field”. In: \textit{Annals of Nuclear Energy} 202 (2024), p. 110491. doi: \url{10.1016/j.anucene.2024.110491}.

[16] Stefan Mihai. “Digital twin: a survey on enabling technologies, challenges, trends and future prospects”. In: \textit{IEEE} 24 (2022), pp. 2255–2289.

[17] I. Pires. Raw dataset with accelerometer, gyroscope and magnetometer data for activities with motion. Available at: \url{https://doi.org/10.17632/XKNHPZ5T96.2}. 2020.

[18] J.Z. Sasiadek and P. Hartana. “Sensor data fusion using Kalman filter”. In: \textit{Proceedings of the Third International Conference on Information Fusion}. Paris, France: IEEE, 2000, WED5/19–WED5/25. doi: \url{10.1109/IFIC.2000.859866}.

[19] A. Singh et al. “Potential applications of digital twin technology in virtual factory”. In: \textit{Digital Twin for Smart Manufacturing}. Ed. by T. Bányai, A. Petrillo, and F. De Felice. Elsevier, 2023, pp. 221–241. doi: \url{10.1016/B978-0-323-99205-3.00011-0}.

[20] Z. Wang. “Digital Twin Technology”. In: \textit{Industry 4.0 - Impact on Intelligent Logistics and Manufacturing}. Ed. by T. Bányai, A. Petrillo, and Fabio De Felice. IntechOpen, 2020. Chap. 1. doi: \url{10.5772/intechopen.80974}.

[21] J. Wu et al. “The Development of Digital Twin Technology Review”. In: \textit{2020 Chinese Automation Congress (CAC)}. Shanghai, China: IEEE, 2020, pp. 4901–4906. doi: \url{10.1109/CAC51589.2020.9327756}.

[22] H. Xiong. “Digital Twin Oriented Visual Saliency Analysis on 360-Degree Panoramic Image”. In: \textit{2022 IEEE 12th International Conference on Electronics Information and Emergency Communication (ICEIEC)}. Beijing, China: IEEE, 2022, pp. 220–223. doi: \url{10.1109/ICEIEC54567.2022.9835071}.

\end{document}